\documentclass[10pt,twocolumn,letterpaper]{article}

\usepackage[applications]{wacv}  
\usepackage[accsupp]{axessibility}
\usepackage{amsmath}
\usepackage{mathrsfs}
\usepackage{amssymb}
\usepackage{subcaption}
\usepackage{graphicx}
\usepackage{algorithm}
\usepackage{algorithmic}
\usepackage{natbib}
\usepackage{lipsum}
\usepackage{pifont}
\usepackage{wasysym}
\usepackage{listings}
\usepackage{tcolorbox}
\usepackage{multirow}
\usepackage{booktabs}
\usepackage{diagbox}
\usepackage{amsmath, amsthm}

\theoremstyle{definition}

\theoremstyle{remark}

\definecolor{wacvblue}{rgb}{0.21,0.49,0.74}
\usepackage[pagebackref,breaklinks,colorlinks,allcolors=wacvblue]{hyperref}

\def\confName{CVPR}
\def\confYear{2026}

\title{3D Cell Oversegmentation Correction via Geo-Wasserstein Divergence}

\author{Peter Chen$^1$ \quad Bryan Chang$^2$ \quad Olivia A Creasey$^3$ \\
Julie Beth Sneddon$^3$ \quad Zev J Gartner$^3$ \quad Yining Liu$^1$\\
$^1$Columbia University \quad $^2$Princeton University \quad $^3$University of California, San Francisco\\
}

\begin{document}
\maketitle
\begin{abstract}
3D cell segmentation methods are often hindered by \emph{oversegmentation}, where a single cell is incorrectly split into multiple fragments. This degrades the final segmentation quality and is notoriously difficult to resolve, as oversegmentation errors often resemble natural gaps between adjacent cells. Our work makes two key contributions. First, we formulate 3D cell oversegmentation as a concrete learning problem and propose a geometry-aware framework to identify and correct these errors. Our approach builds a pre-trained classifier using both 2D geometric and 3D topological features extracted from flawed 3D segmentation results. Second, we introduce a novel metric, Geo-Wasserstein divergence, to quantify changes in 2D geometries. This captures the evolving trends of cell mask shape in a geometry-aware manner. We validate our method through extensive experiments on in-domain plant datasets, including both synthesized and real oversegmented cases, as well as on out-of-domain animal datasets to demonstrate transfer learning performance. An ablation study further highlights the contribution of the Geo-Wasserstein divergence. A clear pipeline is provided for end-users to build pre-trained models to any labeled dataset.
\end{abstract}
    
\vspace{-1em}
\section{Introduction}
\begin{figure}[h!]
  \label{f1}
    \centering
\includegraphics[width=0.75\columnwidth]{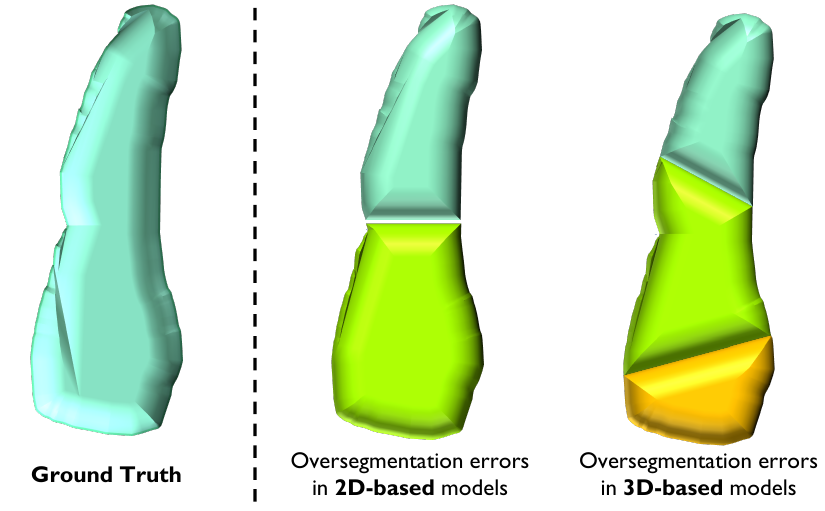} 
    \caption{Examples of oversegmentation errors from 2D-based model (CellStitch) and 3D-based model (PlantSeg).}
    \vspace{-1em}
\end{figure}

3D cell segmentation aims to reconstruct complete 3D cellular structures from sequential 2D microscopy images, a crucial step for downstream analysis of cellular characteristics such as morphology, cell type classification, and volumetric properties \cite{boutros2015,williams2022}. However, noises in raw microscopy images can propagate into the final 3D segmentation, resulting in \emph{oversegmentation}. Shown in Figure \hyperref[f1]{1}, this challenge occurs when individual cells are mistakenly split into multiple segments, which largely downgrades the quality of 3D segmentation results produced the current state-of-the-art methods.

Current 3D cell segmentation approaches can be broadly categorized into \emph{2D-based} and \emph{non-2D-based} methods. Oversegmentation errors could appear differently
depending on which category the segmentation approach
falls under. The 2D-based methods reconstruct the 3D cell structure by ``stitching'' pre-segmented 2D slices layer-by-layer. This strategy has gained popularity recently due to its strong generalizability across diverse cell shapes and datasets, effectively eliminating the need for extensive pre-training on labeled 3D datasets. However, 2D-based approaches heavily rely on the quality of initial 2D segmentation results; thus, errors in individual slices can propagate and accumulate, resulting in oversegmentations predominantly along the imaging axis. In contrast, non-2D-based methods directly reconstruct 3D cell structures from raw image stacks without using 2D pre-segmented results. While this strategy mitigates some errors inherent in 2D segmentation, it requires substantial pre-training on large, dataset-specific annotations, thereby limiting its generalizability. Consequently, oversegmentation errors in non-2D-based methods typically manifest as arbitrary, irregular partitions cutting through the reconstructed cell bodies at random orientations.
\begin{figure*}[t]
    \centering
    \includegraphics[width=0.8\textwidth]{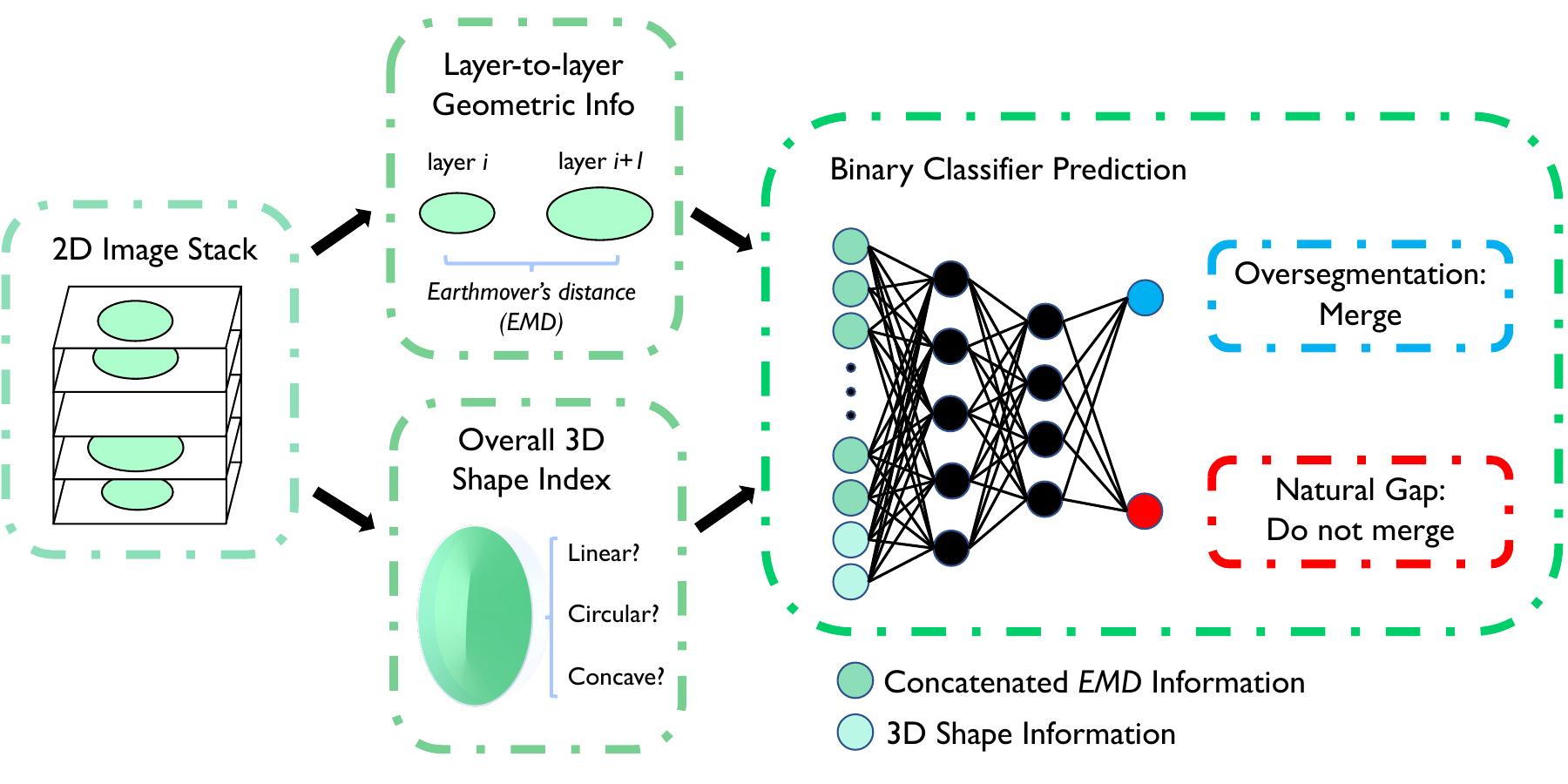} 
    \vspace{-1em}
    \caption{Main pipeline for the oversegmentation correction. The framework extracts layer-to-layer EMDs (Algorithm \hyperref[a2]{2}) and 3D shape information (Algorithm \hyperref[a3]{3}), combining these features for binary classification to distinguish oversegmentations from natural gaps.}
    \label{f2}
    \vspace{-1em}
\end{figure*}

To the best of our knowledge, there is currently no existing literature that explicitly addresses the nature of the oversegmentation problem. This challenge is particularly difficult because oversegmentation errors closely resemble natural gaps separating adjacent cells. In this work, we address the oversegmentation challenge by reformulating it as a binary classification problem, where the learning task is to distinguish oversegmentation from natural gaps between cells. We propose a novel geometric framework that quantitatively captures layer-to-layer 2D geometric consistency as well as overall 3D topological consistency across segmented cell bodies to reliably detect oversegmentation. In addition, we build a pre-trained classifier with the extracted geometric and topological information from flawed 3D segmentation results and then correct these identified oversegmented cell bodies through a proposed interpolation method. Our main contributions are summarized below:

\textbf{(i)} We formally study the oversegmentation issue in 3D cell segmentation as a specific learning problem and introduce a simple yet effective framework for correcting oversegmentation errors. Our framework improves overall 3D segmentation quality and is compatible with various existing state-of-the-art segmentation methods.

\textbf{(ii)} We introduce a novel geometry-aware shape descriptor, \emph{Geo-Wasserstein} divergence, that leverages earthmover’s distance to quantify the divergence between 2D geometric shapes of cell masks. We further highlight its potential application to broader scenarios in general segmentation task for the similar trait of temporal consistency.

\section{Preliminaries and Related Works}

\subsection{3D Cellular Anisotropic Image Segmentation}

The development of cell segmentation has been greatly facilitated by deep learning architectures such as U-Net \cite{ronneberger2015unet}, which have given rise to several state-of-the-art 2D segmentation methods, including CellPose2D \cite{ cellpose2.0, cellpose3.0, cellpose1.0}, Mesmer \cite{greenwald2022wholecell_segmentation}, and StarDist \cite{schmidt2018starconvex}. Building on these advances, 3D segmentation methods can be broadly categorized into two types: \textbf{(i)} 2D-based methods, which typically rely on unsupervised or weakly supervised 2D segmentation followed by subsequent 3D reconstruction, thereby benefiting from reduced complexity and strong generalizability; and \textbf{(ii)} Non-2D-based methods, which directly reconstruct 3D structures from raw image stacks, generally requiring fully supervised training on large labeled datasets. Within it, 2.5D-based approaches have emerged, which integrates the 2D pre-segmented information into the final 3D contextual analysis. Such hybrid framework avoids the need for large-scale 3D pretraining while leveraging the strong generalizability of 2D-based methods. CellPose \cite{cellpose1.0} (2.5D-based), CellStitch \cite{Liu2023Cellstitch} (2D-based), and PlantSeg \cite{plantseg} (non-2D-based) represent the state-of-the-art approaches in their respective categories. However, 3D-based or deep learning-based methods \cite{DL1,DL2,DL3} such as PlantSeg often suffers from limited generalizability due to its reliance on supervised pretraining with large labeled datasets. Technical reviews and baseline choice explanations for more 3D segmentation approaches \cite{wang2022novel, achard2024cellseg3d,DL3,vijayan2024deep}, 2D segmentation backbones \cite{ronneberger2015unet,Graham2019HoverNet,raza2019micro,Chen_2016_CVPR,deng2023omni,yang2020nuset}, and geometry-aware segmentation methods \cite{hu2019topology, pmlr-v202-stucki23a,lux2025topographefficientgraphbasedframework,hu2021topology,Shit_2021} are provided in Appendix \hyperref[aa]{A}.
\vspace{-1em}
\paragraph{CellPose.} CellPose is a 2.5D-based approach that integrates the strengths of both 2D and 3D approaches by leveraging contextual layer awareness to develop a transferable model. It trains models to predict flow vectors for each pixel, and to generate 3D flow vectors, it averages the 2D flow vectors along the $\mathrm{XY}$, $\mathrm{XZ}$, and $\mathrm{YZ}$ planes. However, the substantial difference in sampling ratios between the $\mathrm{XY}$ plane and the $\mathrm{Z}$-axis introduces noise, leading to oversegmentation in highly anisotropic images.
\vspace{-1em}
\paragraph{CellStitch.} CellStitch is an unsupervised 2D-based approach that reconstructs 3D segmentation from CellPose2D's pre-segmentated 2D results. It uses optimal transport to match cells between adjacent layers and introduces an interpolation method to ``stitch'' the discrete, layer-by-layer 2D cell masks into a continuous 3D cell body. However, the performance of such 2D-based methods heavily depends on the quality of the initial 2D segmentation. Specifically, CellPose2D would mis-segment with \hyperref[f3]{empty} or undersegmented mask in 2D layers and causing oversegmentations in CellStitch's 3D results. 
\vspace{-1em}
\paragraph{PlantSeg.} PlantSeg is a supervised non-2D-based method that employs deep learning to predict cell boundaries from 2D image stacks, followed by 3D reconstruction of cell bodies. Its segmentation quality is highly data-driven, depending on the image quality and the similarity between the input images and the training data.

\subsection{Optimal Transport}

Optimal transport (OT) has been applied to mutiple biological areas, including single-cell studies \cite{schiebinger2019optimal}, drug perturbations \cite{bunne2023cellot,chen2024sicnn,chen2025displacement}, and cell alignments \cite{demetci2022scot}. In our framework, OT is utilized to quantify the geometric change between 2D cell masks from adjacent layers.  We introduce OT formulation along with earthmover's distance (EMD) below.

Given measures \( \mathrm{P} \) and \( \mathrm{Q} \) defined on \( \mathbb{R}^d \), along with a transportation cost matrix \(\mathrm{c}(x, y) \), where \( \mathrm{c}(x, y) \) quantifies the transportation cost between a pair of element \(\{ (x,y) \in \{\mathrm{P}\times\mathrm{Q}\}\} \), the objective is to find an optimal transport plan \( \pi \in \Pi(\mathrm{P}, \mathrm{Q}) \) that determines the optimal amount of transportation between pairs \( (x, y) \) which minimizes the overall transportation cost:
\[
\min_{\pi \in \Pi(P, Q)} \sum_{x \in P, y \in Q} \mathrm{c}(x, y) \pi(x, y) \label{e1} \tag{1}.
\]
Formulation \eqref{e1}, known as the Kantorovich formulation of optimal transport, generalizes the Monge formulation \cite{Monge1781} by relaxing the requirement of a direct one-to-one mapping between source and target elements. For two discrete distributions \( X = \{x_i\}_{i=1}^n \) and \( Y = \{y_j\}_{j=1}^m \), the earthmover's distance is defined as the minimum cost required to transform \( X \) into \( Y \):
\vspace{-0.5em}
\begin{align*}
\vspace{-0.5em}
\text{EMD}(X, Y) &= \min_{\pi \in \Pi(X, Y)} \sum_{i=1}^n \sum_{j=1}^m \pi_{ij} \mathrm{c}(x_i, y_j) \label{e2} \tag{2}, \\
\text{s.t.} &\ \ \sum_{j=1}^m \pi_{ij} = x_i, \  \sum_{i=1}^n \pi_{ij} = y_j, \ \pi_{ij} \geq 0, \ \forall i, j.
\vspace{-1em}
\end{align*}
The EMD provides a measurement of the distance between two distributions based on the minimal work required to transform one distribution into the other. Utilizing the EMD, we introduce a novel approach to quantify the divergence between two 2D geometries: \textbf{\emph{Geo-Wasserstein} divergence}. By approximating 2D geometries into discrete distributions of mass (see \S~\hyperref[s3.2]{3.2}), EMD allows us to calculate the minimal transportation cost required to transform one geometry into the other. When cell masks get larger, computing the EMD becomes computationally expensive. Still, \emph{sliced Wasserstein} distance \cite{bonneel2015sliced} can be used a computationally efficient approach to estimate EMD. Typically, increasing the number of projections in sliced Wasserstein method would bring better approximation to the true EMD.

\section{Main Results}
\subsection{Problem Formulation}
\begin{figure}[h!]
    \centering
    \label{f3}
    \includegraphics[width=0.55\columnwidth]{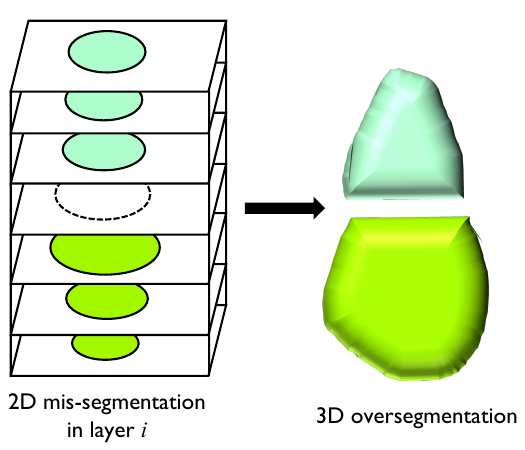} 
    \caption{An example of 2D mis-segmentation from the raw image stack that leads to the subsequent oversegmentation in 3D results.}
\end{figure}
\vspace{-1em}

We first address oversegmentation along the standard axis in 2D- and 2.5D-based models, and then generalize our approach to handle tilted cut planes at arbitrary angles. As illustrated in Figure \hyperref[f3]{3}, when a missing cell slice exists in layer \( i \) (leaving an empty cell mask at layer \( i \)), 2D-based models would incorrectly split the entire cell into two different cells in upper section and lower section. Our task is to: \textbf{(i)} identify a list of suspected oversegmented candidates, consisting of upper and lower parts of cells separated by a gap; \textbf{(ii)} determine whether each candidate represents true oversegmentation or a natural gap, as commonly observed in loosely packed tissues like leaves, where intercellular spaces are frequent; and \textbf{(iii)} recover the correct 2D segmentation by predicting the cell mask for the gap layer and reconstructing the accurate 3D segmentation result. We provide a detailed pipeline in following sections.

\subsection{Method Pipeline}\label{s3.2}
\paragraph{Candidates Screening.} We begin by identifying suspected oversegmented candidates within the flawed 3D segmentation result. As shown in Algorithm \hyperref[a1]{1}, for each cell, we store its mask at the ``top'' (highest layer) and ``bottom'' (lowest layer). We then iterate over all cell pairs in the dataset. If the top-layer mask of one cell overlaps with the bottom-layer mask of another—i.e., they occupy the same position across adjacent layers—and no other cell exists between them, the pair is marked as a suspected oversegmentation candidate. For the suspected candidates collected from candidate screening stage, our goal is to differentiate real oversegmented cases from the case with natural gap between two adjacent cells. Before delving into the technical details of feature extraction, we first introduce some biological background on the \emph{cells used in 3D reconstruction} to provide better context for our method.
\begin{algorithm}[t]\label{a1}
\caption{\textsc{Candidates Screening}}
\begin{algorithmic}[1]
\STATE \textbf{Input:} 3D segmentation result  
\STATE \textbf{Output:} List of oversegmentation candidates pairs

\STATE Initialize \texttt{candidates} = []

\FOR{each pair of cells $(A, B)$ in dataset}
    \IF{highest layer mask of $A$ overlaps with lowest layer mask of $B$}
        \IF{no other complete cell exists in the gap}
            \STATE Append $(A, B)$ to \texttt{candidates}
        \ENDIF
    \ENDIF
\ENDFOR
\end{algorithmic}
\end{algorithm}
\vspace{-0.8em}
\paragraph{Cells used in 3D Segmentation.} 3D segmentation is routinely applied to extract quantitative morphometric information (volumes, neighbour counts, and growth anisotropy) from an entire organ or tissue that has been imaged as a stack. This workflow requires that every voxel from each layer be assigned unambiguously to a single cell, which in practice excludes cells whose highly irregular shapes create multiple, overlapping masks within the same plane. Moreover, standard confocal stacks are anisotropic (i.e., $\Delta x,y\ll\Delta z$). Meanwhile, many irregular cell types (e.g., neuron dendrite) that lie outside compact organs are thinner than the $z$‑sampling interval and appear discontinuous in the volume. Hence, such complex morphologies are usually analyzed with connectomics pipelines based on electron microscopy, rather than with the volumetric segmentation methods for our work. We therefore consider the cells from organ-level image stacks used in 3D reconstruction, assuming a \emph{certain level of regularity} that can be captured through smooth geometric transitions at temporal level. 

Through extensive experiments on plant cells, we validate this assumption in 3D cellular image segmentation. We further show that this pattern generalizes beyond the training distribution: on eleven animal cell stacks, which lack cell walls and therefore exhibit greater irregularity than plant cells, our approach remains effective.

\vspace{-1em}
\paragraph{Geometric Features Extraction.} For each candidate pair, we quantify the geometric change at the gap layer and compare it to the typical layer-to-layer transitions within each cell. In true oversegmentations, the gap-layer change tends to be smooth and consistent with prior transitions. In contrast, an abrupt or irregular change at the gap suggests the presence of two distinct cells, indicating a natural separation. For cell masks \( M \) and \( N \), we represent them as uniform distributions over their geometric regions:
\[
P_M(x) = \frac{1_{\{x \in M\}}}{|M|}, \quad P_N(x) = \frac{1_{\{x \in N\}}}{|N|},
\]
where \( 1_{\{x \in M\}} \) is the indicator function denoting whether \( x \) lies within the boundary of \( M \), and \( |M| \) denotes the area of mask \( M \). The \emph{Geo-Wasserstein} divergence between these two masks—i.e., the change in geometric shape—can then be quantified using the earthmover's distance: \( \text{EMD}(P_M, P_N) \). To assess whether the \emph{Geo-Wasserstein} divergence between two gap-layer masks is consistent with the prior trend of mask transitions in each cell, we use Algorithm \hyperref[a2]{2} to extract EMD values; each cell \( i \) is indexed from layer \( 0 \) (top) to layer \( n_i \) (bottom). 

Moreover, since different cells have varying heights, the lengths of the \(\text{EMD}_{\text{Cell } i}\) arrays differ. To enable training, it is necessary to standardize the input length across all samples. To achieve this, we extract key statistical features from each EMD array to serve as inputs to the model. Specifically, in our pretrained model, we use summary statistics such as the median, max, min, first quartile (\(\mathrm{Q}_{0.25}\)), and third quartile (\(\mathrm{Q}_{0.75}\)) for each array. These statistics are designed to capture higher-level trends in cell shape-change trajectories, which serves as an effective strategy to account for the varying cell heights. We further conduct ablation analysis in \S~\hyperref[s4.4]{4.4} for these statistics to justify the robustness and sensitivity. These features, together with the gap-layer EMD value \(\text{EMD}_\text{gap}\), are used as inputs to the training process.
\begin{algorithm}[t]\label{a2}
\caption{\textsc{Cell EMD Extraction}}
\begin{algorithmic}[1]
\STATE \textbf{Input:} Layer-to-layer masks of Cell A and B
\STATE \textbf{Output:} $\text{EMD}_{\text{Cell A}}$, $\text{EMD}_{\text{gap}}$, $\text{EMD}_{\text{Cell B}}$

\FOR{each layer $i$ to $i+1$ in Cell A}
    \STATE $\text{EMD}_{\text{Cell A}}[i] \gets \text{EMD}(\text{layer } i, \text{layer } i+1)$
\ENDFOR

\STATE $\text{EMD}_{\text{gap}} \gets \text{EMD}(\text{Cell A layer } n_{A}, \text{Cell B layer } 0)$

\FOR{each layer $i$ to $i+1$ in Cell B}
    \STATE $\text{EMD}_{\text{Cell B}}[i] \gets \text{EMD}(\text{layer } i, \text{layer } i+1)$
\ENDFOR
\end{algorithmic}
\end{algorithm}
\vspace{-1em}
\begin{figure}[h]
    \centering
    \label{f4}
    \includegraphics[width=0.6\columnwidth]{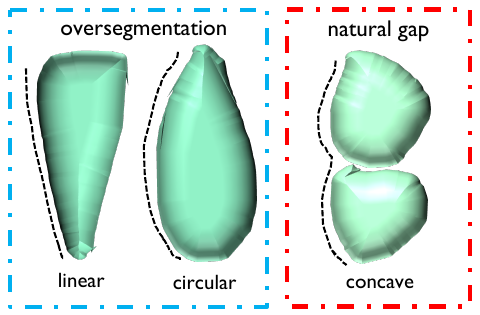} 
    \vspace{-0.5em}
    \caption{Comparison of 3D topological shape between over segmented cases and natural gap cases.}
    \vspace{-2em}
\end{figure}
\paragraph{Topological Shape Extraction.} We propose a straightforward method from 3D topological perspective to distinguish oversegmented cell pairs from those separated by natural gaps. As illustrated in Figure \hyperref[f4]{4}, a key pattern observed across multiple datasets is that oversegmented cases typically exhibit a smooth shape transition at the gap layer, often following either a linear or circular progression. These patterns have biological relevance: circular shapes—wider in the middle and narrower at the ends—are characteristic of mitotic cells undergoing ``circular rounding'' \cite{cr2,cr1}, while gradually tapering linear shapes are typical of structural cells such as palisade cells \cite{pallardy2008physiology}. We quantify such linearity and circularity by analyzing the overlapping area between masks at adjacent layers. For cells exhibiting a linear pattern, the overlap area should change in a monotonically linear trend. While for circular patterns, the overlap area is expected to follow a strictly quadratic trend, characterized by a monotonous increase followed by a decrease. As shown in the Table \hyperref[t1]{1}, we evaluate this strict pattern across cells in the labeled datasets:
\begin{table}[h]
\centering
\label{t1}
\resizebox{\columnwidth}{!}{%
\begin{tabular}{l|cccccc}
\toprule
\textbf{Dataset} & \textbf{Filament} & \textbf{Leaf} & \textbf{Anther} & \textbf{Sepal} & \textbf{Pedicel} & \textbf{Valve} \\
\midrule
\textbf{\# of Cells} & 5,879 & 2,207 & 7,711 & 6,744 & 7,319 & 15,817 \\
\textbf{\% SM}       & 0.963 & 0.927 & 0.920 & 0.908 & 0.869 & 0.851 \\
\bottomrule
\end{tabular}}
\vspace{-0.7em}
\caption{Percentage of cells exhibiting strictly monotonic (\% SM) topological shape pattern across all labeled datasets.}
\end{table}
\vspace{-1em}

However, by examining the cells that deviate from the strictly monotonous pattern, we observed occasional oscillations at specific layers, despite the overall shape following a clear linear or quadratic trend. To improve the inclusiveness towards more candidates, we relaxed the strict constraint by fitting linear and quadratic regressions to the overlap curve and using the normalized regression coefficient \( \widetilde{\mathcal{R}^2} \) as a measure of shape consistency. For strictly monotonous cells, we assign \( \mathcal{R}^2 = 1 \). Each suspected candidate pair is treated as a single cell, and we apply Algorithm~\hyperref[a3]{3} to extract its shape information.

\begin{algorithm}[h]\label{a3}
\caption{\textsc{Topological Shape Classification}}
\begin{algorithmic}[1]
\STATE \textbf{Input:} Integrated cell masks from A and B
\STATE \textbf{Output:} Class (linear or quadratic), ShapeIndex ($\widetilde{\mathcal{R}^2}$)

\FOR{each layer $i$ to $i+1$ in Cell}
    \STATE $\text{OverlapArea}[i] \gets \text{Overlap}(\text{layer } i, \text{layer } i+1)$
\ENDFOR

\STATE $\text{Quadratic } \mathcal{R}^2 \gets \text{QuadraticRegression}(\text{OverlapArea})$
\STATE $\text{Linear }\mathcal{R}^2 \gets \text{LinearRegression}(\text{OverlapArea})$
\STATE $\mathcal{R}^2 \gets \max(\text{Linear }\mathcal{R}^2, \text{Quadratic }\mathcal{R}^2)$
\STATE $\text{Class} \gets \arg\max(\text{Linear }\mathcal{R}^2, \text{Quadratic } \mathcal{R}^2)$

\IF{$\text{OverlapArea}$ is monotonous}
    \STATE $\mathcal{R}^2 \gets 1$
\ENDIF
\STATE \text{ShapeIndex} $\widetilde{\mathcal{R}^2}$ $\gets$ Normalized $\mathcal{R}^2$ from each Class
\end{algorithmic}
\end{algorithm}

\vspace{-1.5em}

\paragraph{Classifier Training.} By leveraging both sources of information, we aim to evaluate whether the gap layer maintains geometric consistency with the preceding layers, and whether the entire cell exhibits topological consistency typical of normal cells. Our full pipeline for classifier training is illustrated in Figure~\hyperref[f2]{2}. We begin by constructing a training set composed of labeled true (oversegmented) and false (natural gap) cases from annotated datasets. True cases are generated by simulating oversegmentation—removing a specific layer from a 2D segmentation to create an artificial gap. False cases are obtained by applying Algorithm~\hyperref[a1]{1} to identify naturally separated cells. We then apply Algorithm~\hyperref[a2]{2} and Algorithm~\hyperref[a3]{3} to each candidate in the training set to extract geometric and topological shape features. To preserve all extracted information, we concatenate the features and use them as input to a multi-layer perceptron, which is trained to perform binary classification.
\vspace{-1em}
\paragraph{Cross-layer Interpolation.} For final oversegmented cell pairs, we recover the mis-segmented 2D mask between them (at layer \( i \)) by interpolating between the cell mask boundaries at layers \( i-1 \) and \( i+1 \), leveraging the interpolation methods \cite{Liu2023Cellstitch,Solomon2015}. The corrected 2D segmentation masks are then used to reconstruct the updated 3D segmentation. The technical details of cross-layer interpolation are provided in Appendix B.1.
\vspace{-1em}
\paragraph{Computational Efficiency.} We highlight the computational efficiency of this lightweight, effective method and report timings from real applications. Processing a large $300 \times 2008 \times 2008\ (\mathrm{Z}\times\mathrm{X}\times\mathrm{Y})$ animal stack takes $\sim$1 hour on two NVIDIA A40 GPUs (48GB each). On a medium $2000 \times 224 \times 224$ plant dataset, it takes $\sim$7 minutes to identify oversegmentation candidates and $\sim$15 minutes to stitch and recover all oversegmented cells, using CPUs only.

\begin{figure*}[t]
    \centering
\includegraphics[width=0.8\textwidth]{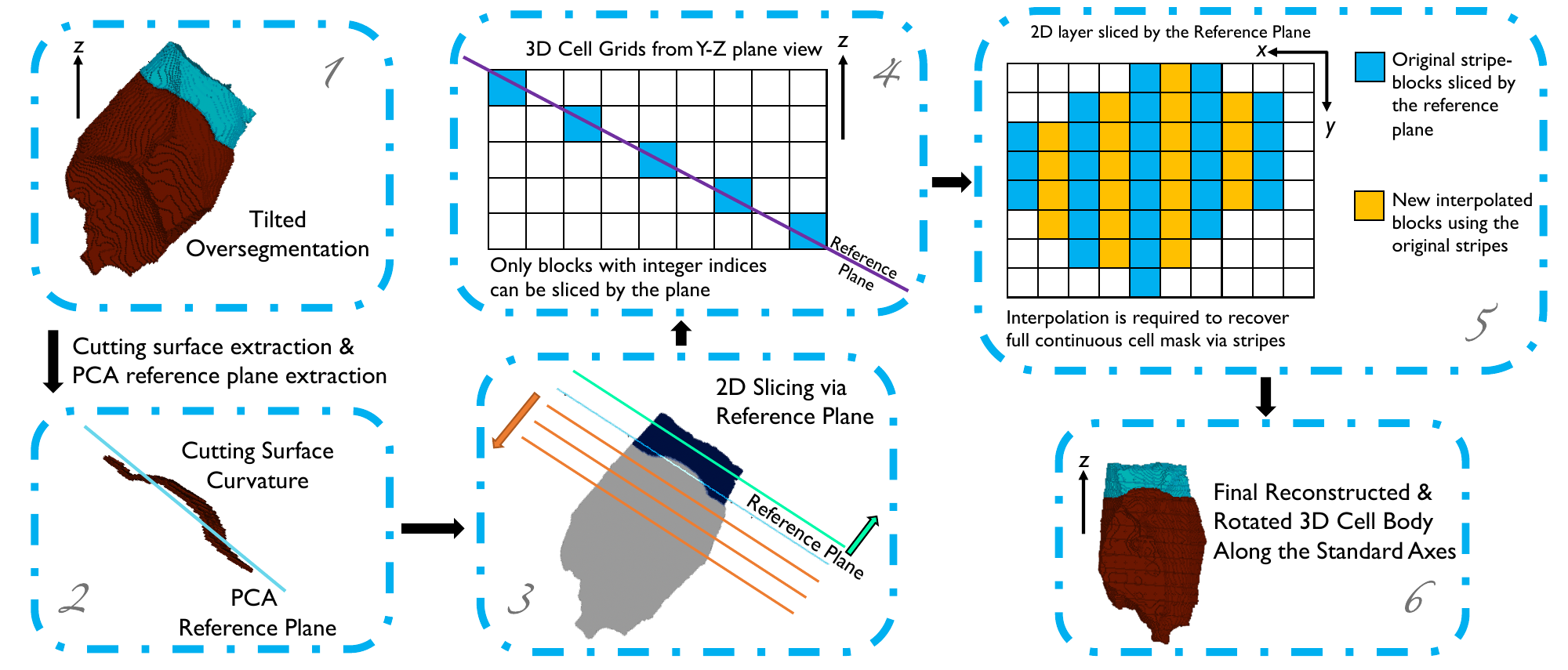} 
    \vspace{-0.5em}
    \caption{Main pipeline for constructing rotated 3D segmentation layers for the tilted cases from 3D-based method (e.g., PlantSeg).}
    \label{f5}
    \vspace{-1.2em}
\end{figure*}

\subsection{Method Generalization}\label{s3.3}
Recall that our original framework operates on oversegmentation results where the cutting surfaces between cell pairs are always aligned with the horizontal planes of the standard coordinate axes. In contrast, non-2D-based approaches often produce cutting surfaces that are tilted, irregular, or rugged. To address this discrepancy, we extend our framework to accommodate these more complex oversegmentation cases, illustrated in Figure~\hyperref[f5]{5}.
\vspace{-1em}
\paragraph{Candidates Screening.} Since oversegmentation can occur in any direction for 3D-based approaches, we store all cell information in a graph structure and retrieve neighboring cells (i.e., those in direct contact) as suspected candidates. The number of suspected candidates in this setting is expected to be significantly larger than in the axis-aligned case described in \S~\hyperref[s3.2]{3.2}.
\vspace{-1em}
\paragraph{Oversegmentation Cutting Surface.} We define the cutting surface between two cells as the shared curved interface where the cells are in direct contact. In the previous setting, this surface was aligned with the $\mathrm{XY}$ plane, and 2D segmentation layers were sliced from the 3D volume along this standard plane. In this generalized case, we extend this concept by slicing the 3D segmentation volume layer-by-layer along a rotated reference plane (one that is aligned with the cutting surface between two oversegmented cells) thereby forming a rotated 2D segmentation representation.

\vspace{-1em}
\paragraph{Rotated 3D Segmentation Reconstruction.} The first step is to estimate a reference plane from the rugged cutting surface between two oversegmented cells. We treat this cutting surface as a curvature and apply PCA to extract a best-fit plane, which serves as the \emph{estimated reference plane}. Using this reference plane, we slice the 3D segmentation volume into parallel 2D layers by shifting the plane upward and downward. However, as illustrated in Step 4 of Figure~\hyperref[f5]{5}, since all cells are stored in a discrete index format, the tilted plane intersects each layer as a series of discrete strips (aligned with integer indices), rather than forming smooth, continuous 2D geometries. To address this, we interpolate between these strips to reconstruct complete 2D cell masks. The reconstructed 2D masks are then used as input to our pretrained classifier. Specifically, we remove the masks of both cells at the reference layer (i.e., the unshifted PCA plane) and treat the remaining upward and downward segments as a pair of candidate cell fragments. 

The classifier then determines whether these two parts should be merged. Notably, in this setting, layer-to-layer interpolation is no longer required to correct 2D mis-segmentation—if merging is validated, we directly combine the original 3D cell bodies. We evaluate these cases specifically for PlantSeg. Results are shown in Table \hyperref[t2]{2(a)}.

\section{Experiments}

\paragraph{Dataset Overview.} We evaluate our method on two types of datasets: publicly available, labeled plant cell datasets, and private, unlabeled animal cell datasets. Specifically, the animal datasets \textbf{lack ground-truth annotation}. In total, we use six plant cell datasets \cite{bassel2019_dataset}, each consisting of 100 image stacks, and an animal cell datasets with 11 image stacks. Notably, each animal image stack is a large volumetric dataset containing approximately 300 image layers, with the maximum \(\mathrm{X} \times \mathrm{Y}\) plane size reaching up to \(2008 \times 2008\) pixels. After the initial 3D segmentation, each image stack contains approximately 2,000 to 4,000 cells. Note that due to the scarcity of high-quality and labeled 3D volumetric image stacks, plant datasets from \cite{bassel2019_dataset} are commonly used as a standard benchmark for 3D cell segmentation training, as adopted by state-of-the-art methods \cite{Liu2023Cellstitch,eschweiler2022robust3dcellsegmentation,plantseg,DL2}. To demonstrate the effectiveness under transfer learning, we additionally evaluated on our private animal cell datasets without re-training, showing its applicability across different cell types in organ-level image stacks. Further dataset details are provided in Appendix D.
\vspace{-1em}
\paragraph{Experiment Overview.} Our experimental evaluation consists of four parts: \textbf{(i)} synthesized test cases derived from ground-truth plant datasets, which provide deterministic labels and enable accurate evaluation using metrics such as recall and F1-score; \textbf{(ii)} direct augmentation of real, flawed 3D plant segmentation results, which we evaluate it across multiple state-of-the-art methods using both deterministic quantitative metrics and qualitative results validated by human experts; \textbf{(iii)} direct augmentation of flawed 3D segmentation results on animal datasets, which serves to assess the transferability of our method in the absence of ground truth; and \textbf{(iv)} ablation analysis on the key design: the robustness of the statistical feature extraction and the contribution brought by the \emph{Geo-Wasserstein} divergence.

\begin{table*}[t]
\centering
\label{t2}

\begin{subtable}[t]{0.65\textwidth}
\centering
\caption{Results on real oversegmented cases produced by different 3D methods from plant cell dataset}
\resizebox{\linewidth}{!}{%
\begin{tabular}{l|cccccc}
\toprule
& \textbf{Leaf} & \textbf{Sepal} & \textbf{Valve} & \textbf{Pedicel} & \textbf{Anther} & \textbf{Filament} \\
\midrule
\multicolumn{7}{c}{\textbf{\textsc{CellPose}-based Results}} \\
\midrule
\textsc{\textbf{mAP}} — \textsc{CellPose}         & 0.476$\pm$0.171 & 0.428$\pm$0.133 & 0.270$\pm$0.091 & 0.358$\pm$0.143 & 0.383$\pm$0.154 & 0.473$\pm$0.172 \\
\textsc{\textbf{mAP}} — \textsc{CellPose$+$Ours}  & 0.514$\pm$0.179 & 0.447$\pm$0.136 & 0.286$\pm$0.091 & 0.388$\pm$0.146 & 0.403$\pm$0.151 & 0.502$\pm$0.178 \\
\midrule
\textsc{\textbf{Jac}} — \textsc{CellPose}          & 0.601$\pm$0.196 & 0.611$\pm$0.155 & 0.549$\pm$0.089 & 0.497$\pm$0.174 & 0.637$\pm$0.146 & 0.646$\pm$0.190 \\
\textsc{\textbf{Jac}} — \textsc{CellPose$+$Ours}  & 0.655$\pm$0.179 & 0.640$\pm$0.149 & 0.568$\pm$0.087 & 0.531$\pm$0.166 & 0.664$\pm$0.143 & 0.704$\pm$0.178 \\
\midrule
(\textsc{Correct}, \textsc{Unsure}, \textsc{Incorrect}) & (35, 4, 5) & (142, 59, 22) & (249, 81, 43) & (136, 36, 23) & (139, 53, 34) & (100, 18, 25) \\
\textbf{\textsc{Case-by-case Accuracy}}   & 0.875 & 0.866 & 0.853 & 0.855 & 0.803 & 0.800 \\
\midrule
\multicolumn{7}{c}{\textbf{\textsc{CellStitch}-based Results}} \\
\midrule
\textsc{\textbf{mAP}} — \textsc{CellStitch}  & 0.581$\pm$0.140 & 0.268$\pm$0.124 & 0.457$\pm$0.099 & 0.241$\pm$0.181 & 0.422$\pm$0.122 & 0.402$\pm$0.161 \\
\textsc{\textbf{mAP}} — \textsc{CellStitch}$+$\textsc{Ours} & 0.606$\pm$0.148 & 0.275$\pm$0.125 & 0.471$\pm$0.100 & 0.247$\pm$0.182 & 0.438$\pm$0.126 & 0.410$\pm$0.156 \\
\midrule
\textsc{\textbf{Jac}} — \textsc{CellStitch}         & 0.634$\pm$0.136 & 0.443$\pm$0.154 & 0.720$\pm$0.075 & 0.361$\pm$0.154 & 0.611$\pm$0.163 & 0.495$\pm$0.137 \\
\textsc{\textbf{Jac}} — \textsc{CellStitch}$+$\textsc{Ours}  & 0.668$\pm$0.133 & 0.456$\pm$0.153 & 0.737$\pm$0.075 & 0.371$\pm$0.155 & 0.620$\pm$0.165 & 0.504$\pm$0.133 \\
\midrule
(\textsc{Correct}, \textsc{Unsure}, \textsc{Incorrect}) & (42, 38, 2) & (70, 113, 13) & (75, 82, 4) & (71, 134, 15) & (93, 141, 13) & (92, 124, 28) \\
\textbf{\textsc{Case-by-case Accuracy}}    & 0.933 & 0.843 & 0.949 & 0.826 & 0.881 & 0.767 \\
\midrule
\multicolumn{7}{c}{\textbf{\textsc{PlantSeg}-based Results}} \\
\midrule
\textsc{\textbf{mAP}} — \textsc{PlantSeg}         & 0.564$\pm$0.081 & 0.377$\pm$0.131  & 0.547$\pm$0.098  & 0.607$\pm$0.160 & 0.563$\pm$0.100  & 0.484$\pm$0.098 \\
\textsc{\textbf{mAP}} — \textsc{PlantSeg$+$Ours}  & 0.581$\pm$0.082 & 0.391$\pm$0.132 & 0.560$\pm$0.098 & 0.621$\pm$0.155 & 0.579$\pm$0.108 & 0.505$\pm$0.097 \\
\midrule
\textsc{\textbf{Jac}} — \textsc{PlantSeg}         & 0.692$\pm$0.110 & 0.541$\pm$0.116  & 0.635$\pm$0.079 & 0.887$\pm$0.053  & 0.748$\pm$0.168  & 0.668$\pm$0.102 \\
\textsc{\textbf{Jac}} — \textsc{PlantSeg$+$Ours}  & 0.709$\pm$0.109 & 0.567$\pm$0.123 & 0.663$\pm$0.080 & 0.890$\pm$0.052  & 0.750$\pm$0.167 & 0.678$\pm$0.112  \\
\midrule
(\textsc{Correct}, \textsc{Unsure}, \textsc{Incorrect}) & (19, 1, 1) & (40, 2, 3)  & (125, 26, 12)  & (84, 12, 11)  & (77, 8, 16) & (90, 11, 6)  \\
\textbf{\textsc{Case-by-case Accuracy}}    & 0.950 & 0.930 & 0.912 & 0.884 &  0.828 & 0.901 \\
\bottomrule
\end{tabular}%
}
\end{subtable}
\hfill
\begin{subtable}[t]{0.34\textwidth}
\centering
\caption{Results on correcting synthesized oversegmented cases from labeled plant cell dataset}
\resizebox{\linewidth}{!}{%
\begin{tabular}{l|cccccc}
\toprule
\textbf{Dataset}    & \textbf{Sepal} & \textbf{Pedicel} & \textbf{Valve} & \textbf{Leaf} & \textbf{Anther} & \textbf{Filament} \\
\midrule
\textbf{Recall}    & 0.955 & 0.854 & 0.965 & 0.988 & 0.840 & 0.854 \\
\textbf{Precision} & 0.963 & 0.998 & 0.983 & 1.000 & 0.985 & 0.981 \\
\textbf{F1-Score}  & 0.959 & 0.920 & 0.974 & 0.994 & 0.907 & 0.913 \\
\bottomrule
\end{tabular}%
}
\vspace{1.5em}
\centering
\caption{Results on correcting real oversegmented cases produced by 3D method from animal pancreas (PA) dataset; Given the lack of ground truth labeling, only qualitative metric is reported}
\resizebox{\linewidth}{!}{%
\begin{tabular}{l|cccccc}
\toprule
\textbf{Dataset} & \textbf{PA-11} & \textbf{PA-16} & \textbf{PA-03} & \textbf{PA-07} & \textbf{PA-04} & \textbf{PA-09} \\
\midrule
\textbf{\# \textsc{Correct}}   & 9  & 10 & 19 & 29 & 21 & 25 \\
\textbf{\# \textsc{Unsure}}    & 1  & 3  & 3  & 5  & 3  & 9  \\
\textbf{\# \textsc{Incorrect}} & 1  & 2  & 5  & 4  & 3  & 6  \\
\midrule
\textbf{\textsc{Accuracy}}     & 0.900 & 0.833 & 0.792 & 0.879 & 0.875 & 0.807 \\
\bottomrule
\end{tabular}%
}

\vspace{0.81em}

\resizebox{\linewidth}{!}{%
\begin{tabular}{l|cccccc}
\toprule
\textbf{Dataset} & \textbf{PA-10} & \textbf{PA-05} & \textbf{PA-02} & \textbf{PA-14} & \textbf{PA-08} & \textbf{Overall} \\
\midrule
\textbf{\# \textsc{Correct}}   & 35 & 52 & 28 & 52 & 94 & 374 \\
\textbf{\# \textsc{Unsure}}    & 8  & 7  & 8  & 21 & 24 & 92  \\
\textbf{\# \textsc{Incorrect}} & 4  & 10 & 1  & 2  & 14 & 52  \\
\midrule
\textbf{\textsc{Accuracy}}     & 0.897 & 0.839 & 0.966 & 0.963 & 0.870 & 0.878 \\
\bottomrule
\end{tabular}%
}
\end{subtable}
\vspace{-0.5em}
\caption{\textbf{(a)} We report mean Average Precision (\textsc{\textbf{mAP}}) and Jaccard index (\textsc{\textbf{Jac}}) under CellPose, CellStitch and PlantSeg, with and without our method (``$+$Ours''). The last two rows in each method show case-by-case human verification results. Valid correction are the sum of correct and incorrect cases, and the final accuracy is the ratio of correct cases to valid fixations; \textbf{(b)} Results on synthesized plant cases; \textbf{(c)} Results on animal dataset, with accuracy calculated in the same way as in (a).}
\vspace{-1.5em}
\end{table*}

\subsection{Plant Cells: Synthesized Cases}

In this task, we reserve a small portion (10 stacks) from each plant dataset, containing true (oversegmented) and false (natural gap) test cases that are excluded from the training set. This setup allows us to evaluate the pretrained model on unseen but similar data. Since these test cases are constructed from labeled ground-truth segmentations, each has a deterministic ground-truth label, enabling precise quantitative evaluation. Results are shown in Table \hyperref[t2]{2(b)}.

\subsection{Plant Cells: Real Cases}\label{s4.2}

In this task, we apply CellPose, CellStitch, and PlantSeg directly to the raw microscopy images to obtain the initial 3D segmentation results, which inherently contain errors due to imperfect segmented cell masks, such as shrinkage and fragmentation, commonly encountered in real-world applications. We then apply our method across \S~\hyperref[s3.2]{3.2} and  \S~\hyperref[s3.3]{3.3} to these flawed segmentations. This experiment is designed to evaluate the robustness of our framework in realistic segmentation settings with errors from pre-2D results. 
\vspace{-1em}
\paragraph{Evaluation Metrics.} We first report the quantitative results to directly assess the improvement in overall segmentation quality using standard evaluation metrics from multiple perspectives: mean Average Precision (mAP) and Jaccard Index (Jac). Specifically, the average precision (AP) is computed as $\mathrm{TP} / (\mathrm{TP} + \mathrm{FN} + \mathrm{FP})$, where $\mathrm{TP}$, $\mathrm{FN}$, and $\mathrm{FP}$ represent the numbers of true positive, false negative, and false positive masks, respectively, under a specified Intersection over Union (IoU) threshold $t$. The mean Average Precision (mAP) is then obtained by averaging the AP scores at IoU thresholds of $t = \{0.25,0.5, 0.75\}$. The Jaccard Index (Jac), by contrast, directly measures the IoU overlap between predicted and ground truth cell masks. Results are shown in Table \hyperref[t2]{2(a)}. 

\begin{figure}[t]
    \centering
\includegraphics[width=0.85\columnwidth]{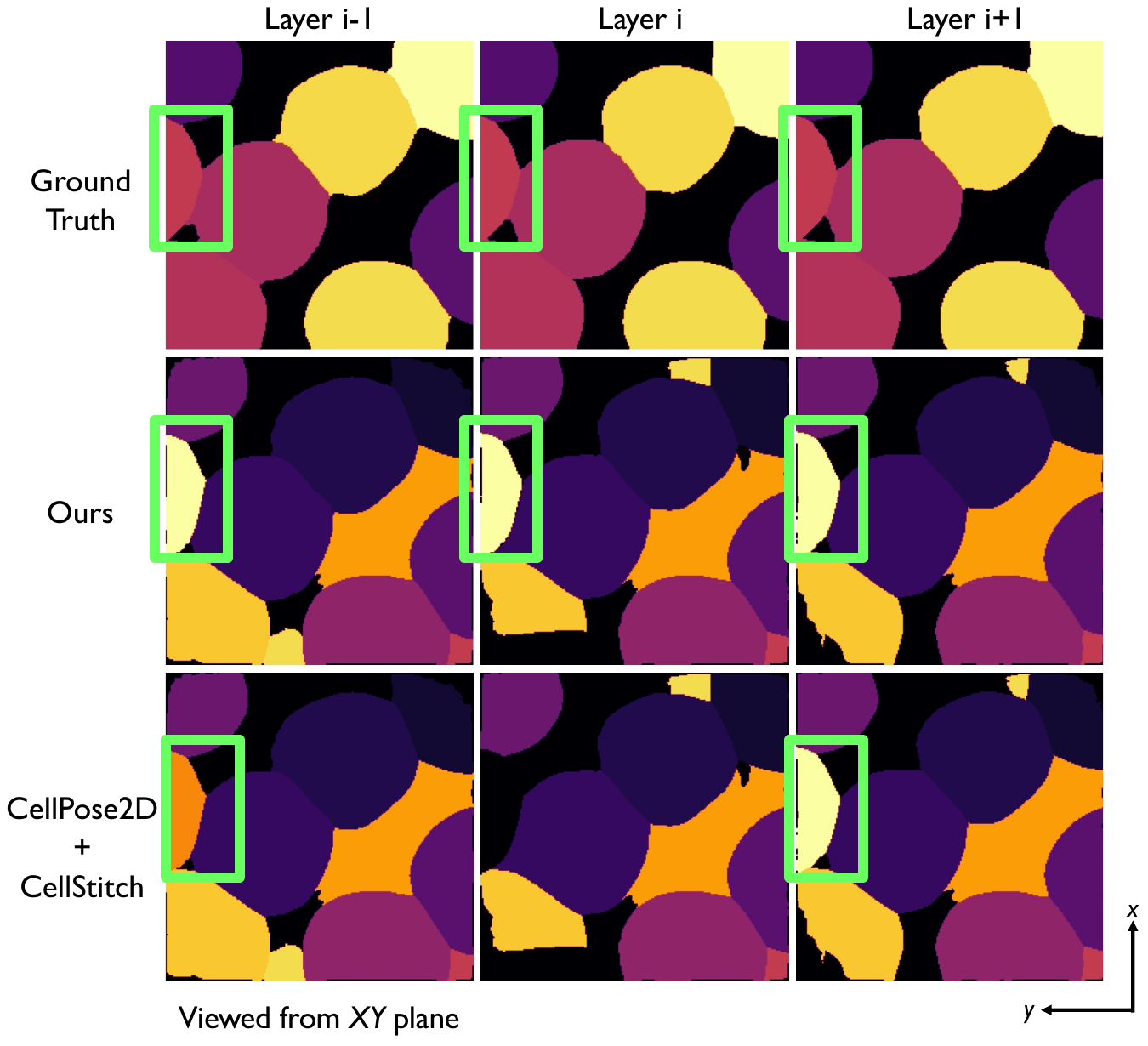} 
\vspace{-0.5em}
    \caption{Example of correct case. In the CellStitch row, layers $i-1$ and $i+1$ contain cell masks (highlighted using the green box) that are also appeared in the ground truth row at the same position. Our method detects and stitches them as the same cell.}
    \label{f6}
    \vspace{-2em}
\end{figure}

Apart from quantitative results that reflects the overall segmentation quality improvement, we further incorporate case-by-case validation from human expert to get a deeper understanding of the specific correction quality. During human verification, we classify oversegmentation corrections into three categories: \textbf{(i)} Correct cases, where the oversegmented cell masks are present in the ground truth and belong to the same cell; \textbf{(ii)} Incorrect cases, where the oversegmented cell masks are present in the ground truth but actually belong to different cells (natural gap); and \textbf{(iii)} Unsure cases, which include ambiguous scenarios such as ``hallucination'', where the correction appears geometrically plausible but lacks definitive support from the ground truth. We provide further discussion and visualization examples for these cases in Appendix C. An example of the correct case is shown in Figure \hyperref[f6]{6}, where in the CellStitch row, layers $i-1$ and $i+1$ contain cell masks (highlighted using the green box) that are also appeared in the ground truth row at the same position. Our method detects these two masks and stitches them as the same cell. 

\subsection{Animal Cells: Real Cases}

In this task, we evaluate the transfer learning effectiveness of our pre-trained plant cell model on an unlabeled animal pancreas cell dataset. Due to the lack of ground truth, we present the augmented results on CellStitch and report case-by-case verification result. Results are shown in Table \hyperref[t2]{2(c)}.

Together with the results in \S~\hyperref[s4.2]{4.2}, we observe that the performance of our framework is the following source: CellPose2D often shrinks the cell mask smaller during 2D segmentation, especially when the cell boundaries in the raw images are blurred by a light band. This shrinkage results in the loss of geometric information, as the EMD between the two masks is also affected when one mask shrinks. Moreover, the final result is also influenced by the resolution of the image stack and the average size of the cell masks. Higher resolutions amplify geometric details, enabling \emph{Geo-Wasserstein} divergence to capture more nuanced information.

\subsection{Ablation Analysis}\label{s4.4}
\vspace{-1em}
\begin{table}[h]\label{t3}
\centering
\setlength{\tabcolsep}{6pt}
\renewcommand{\arraystretch}{1.1}
\resizebox{0.95\linewidth}{!}{%
\begin{tabular}{l|cc|cc|cc}
\toprule
\textbf{Metric} & \multicolumn{2}{c|}{\textbf{Leaf}} & \multicolumn{2}{c|}{\textbf{Sepal}} & \multicolumn{2}{c}{\textbf{Pedicel}} \\
\midrule
& w/o EMD & w EMD & w/o EMD & w EMD & w/o EMD & w EMD \\
\midrule
\textbf{\# \textsc{Correct}}   & 52 & 42 & 126 & 70  & 78  & 71 \\
\textbf{\# \textsc{Unsure}}    & 64 & 38 & 312 & 113 & 477 & 134 \\
\textbf{\# \textsc{Incorrect}} & 36 & 3  & 131 & 13  & 128 & 15 \\
\midrule
\textbf{\textsc{Accuracy}}     & 0.5909 & \textbf{0.9333} & 0.4903 & \textbf{0.8434} & 0.3786 & \textbf{0.8256} \\
\bottomrule
\end{tabular}
}

\resizebox{0.95\linewidth}{!}{%
\begin{tabular}{l|cc|cc|cc}
\toprule
\textbf{Metric} & \multicolumn{2}{c|}{\textbf{Anther}} & \multicolumn{2}{c|}{\textbf{Filament}} & \multicolumn{2}{c}{\textbf{Valve}} \\
\midrule
& w/o EMD & w EMD & w/o EMD & w EMD & w/o EMD & w EMD \\
\midrule
\textbf{\# \textsc{Correct}}   & 113 & 93 & 139 & 92 & 91 & 75 \\
\textbf{\# \textsc{Unsure}}    & 371 & 141 & 338 & 124 & 148 & 82 \\
\textbf{\# \textsc{Incorrect}} & 102 & 13  & 99  & 28  & 40  & 4 \\
\midrule
\textbf{\textsc{Accuracy}}     & 0.5256 & \textbf{0.8807} & 0.5840 & \textbf{0.7667} & 0.6945 & \textbf{0.9494} \\
\bottomrule
\end{tabular}
}
\vspace{-0.5em}
\caption{Results on the ablation. ``w/o EMD'' and ``w EMD'' represent the result without and with using EMD, respectively.}
\vspace{-2.5em}
\end{table}

\paragraph{\emph{Geo-Wasserstein} Divergence.} We train the classifier by discarding the \emph{Geo-Wasserstein} features and evaluate on CellStitch’s flawed 3D segmentation results for labeled plant cells. Case-by-case verification results are reported in Table \hyperref[t3]{3}. Incorporating EMD significantly improved correction accuracy by capturing geometric differences between 2D masks that area-based metrics often overlook. While 3D topological features, such as layer-to-layer overlaps, offer limited information, geometry-aware EMD reliably detects subtle layer-to-layer 2D mask shape variations critical for distinguishing true oversegmentations from natural gaps. For instance, two geometries may have identical areas but entirely different shapes, then \emph{Geo-Wasserstein} could capture this shape divergence, while the other area-based metrics failed to do so.
\vspace{-1em}
\paragraph{Statistical Features.} Following \S~\hyperref[s3.2]{3.2}, for each candidate gap layer~\(i\) we summarize the layer-to-layer EMD trajectories on both sides of the gap. For each side we compute \(\{\min,\max,\mathrm{median},\mathrm{Q}_{0.25},\mathrm{Q}_{0.75}\}\) of the per-layer EMDs. The range \((\min,\max)\) encodes largest shift, while \((\mathrm{median},\mathrm{Q}_{0.25},\mathrm{Q}_{0.75})\) captures the internal trend, enabling us to assess whether the two fragments exhibit compatible statistics consistent with a single connected cell despite varying cell heights. 

We evaluate robustness and sensitivity with three variants: \textbf{(i) Incomplete}: use only \(\{\mathrm{median},\mathrm{Q}_{0.25},\mathrm{Q}_{0.75}\}\) to test whether range features are necessary;
\textbf{(ii) Extra}: augment with deciles \(\{\mathrm{Q}_{0.1},\mathrm{Q}_{0.2},\ldots,\mathrm{Q}_{0.9}\}\) plus \(\min\) and \(\max\) to test whether denser quantiles better capture trends;
\textbf{(iii) Perturbed}: replace \(\{\min,\mathrm{median},\max,\mathrm{Q}_{0.25},\mathrm{Q}_{0.75}\}\) with the five quantiles \(\{\mathrm{Q}_{0.1},\mathrm{Q}_{0.3},\mathrm{Q}_{0.5},\mathrm{Q}_{0.7},\mathrm{Q}_{0.9}\}\), keeping dimensionality fixed to probe sensitivity to the choice of quantile levels. Results are shown in Table \hyperref[t4]{4}:
\vspace{-0.5em}
\begin{table}[h]
\centering
\label{t4}
\setlength{\tabcolsep}{3.5pt}
\renewcommand{\arraystretch}{1.05}
\resizebox{\columnwidth}{!}{
\begin{tabular}{l|cccccc}
\toprule
 & \textbf{Leaf} & \textbf{Sepal} & \textbf{Valve} & \textbf{Pedicel} & \textbf{Anther} & \textbf{Filament} \\
\midrule
\multicolumn{7}{c}{\textbf{\textsc{Statistical Feature Variants}}} \\
\midrule
\textsc{\textbf{mAP}} — \textsc{Incomplete} & $0.481\pm0.174$ & $0.407\pm0.141$ & $0.255\pm0.093$ & $0.364\pm0.148$ & $0.380\pm0.157$ & $0.461\pm0.176$ \\
\textsc{\textbf{mAP}} — \textsc{Extra}      & $0.516\pm0.178$ & $0.454\pm0.153$ & $0.272\pm0.124$ & $0.404\pm0.144$ & $0.407\pm0.152$ & $0.499\pm0.173$ \\
\textsc{\textbf{mAP}} — \textsc{Perturbed}  & $0.510\pm0.175$ & $0.444\pm0.137$ & $0.289\pm0.099$ & $0.390\pm0.145$ & $0.401\pm0.154$ & $0.497\pm0.172$ \\
\midrule
\textsc{\textbf{Jac}} — \textsc{Incomplete} & $0.611\pm0.201$ & $0.602\pm0.154$ & $0.550\pm0.091$ & $0.516\pm0.178$ & $0.636\pm0.147$ & $0.637\pm0.182$ \\
\textsc{\textbf{Jac}} — \textsc{Extra}      & $0.660\pm0.177$ & $0.638\pm0.162$ & $0.555\pm0.103$ & $0.542\pm0.168$ & $0.661\pm0.144$ & $0.691\pm0.171$ \\
\textsc{\textbf{Jac}} — \textsc{Perturbed}  & $0.652\pm0.168$ & $0.636\pm0.147$ & $0.564\pm0.088$ & $0.530\pm0.166$ & $0.663\pm0.145$ & $0.693\pm0.176$ \\
\bottomrule
\end{tabular}
\vspace{-1em}
}

\caption{Ablation on statistical feature variants.}

\end{table}
\vspace{-1em}

We find that incomplete range information (i) markedly degrades performance, whereas adding extra features (ii) yields no significant improvement over the original setup and can introduce noise. The original configuration is also robust to perturbations (iii). The number of oversegmentation candidates is comparable for (ii), (iii), and the original setup, while (i) identifies noticeably more cell-pair candidates due to less restrictions from the removed features.
\vspace{-0.5em}
\section{Conclusion}

We present a geometric framework for 3D cell oversegmentation, an important yet unresolved challenge that degrades the quality of 3D segmentation results. Beyond its impact on 3D cell segmentation, the proposed \emph{Geo-Wasserstein} divergence is broadly applicable to segmentation problems, especially video segmentation, where temporal consistency mirrors the slice-to-slice geometric consistency exploited in 3D cell stacks: in both settings, the key question is whether regions in neighboring frames (time) or slices (depth) correspond to the same object. We hope this framework facilitates adoption by end-users and inspires the vision and geometry processing community to build upon it.

~\

\vspace{-0.5em}
\noindent\textbf{Acknowledgment.} We acknowledge computing resources from Columbia University’s Shared Research Computing Facility project, which is supported by NIH Research Facility Improvement Grant 1G20RR030893-01, and associated funds from the New York State Empire State Development, Division of Science Technology and Innovation (NYSTAR) Contract C090171, both awarded April 15, 2010.


{
    \small
    \bibliographystyle{ieeenat_fullname}
    \bibliography{main}
}

\newpage

\appendix
\onecolumn
\section{Further Related Works}

In this section, we provide further discussion of the 2D segmentation techniques used in CellPose and CellStitch, as well as some alternative 3D segmentation approaches and their limitations when applying to our focused task.
\vspace{-1em}
\paragraph{2D Segmentation Methods.} We introduce the following key 2D segmentation methods that facilitate the development of 3D cell segmentation methods:

U-Net \cite{ronneberger2015unet} is a fully convolutional neural network specifically developed for biomedical image segmentation. It employs an encoder-decoder structure with skip connections that preserve spatial information, making it ideal for cell and tissue segmentation. Hover-Net \cite{Graham2019HoverNet} combines semantic segmentation with instance-level clustering via predicted horizontal and vertical distance maps (hover maps). Hover-Net is particularly effective for segmenting densely packed cell nuclei in pathology slides.  Micro-Net \cite{raza2019micro} is a lightweight CNN architecture specifically tailored for segmenting cells in resource-limited environments. It leverages depth-wise separable convolutions and inverted residuals to achieve efficient segmentation with fewer parameters. DCAN \cite{Chen_2016_CVPR} simultaneously predicts pixel-wise object masks and object contours using a dual-branch CNN architecture. Effective for precisely delineating individual cells by explicitly modeling boundaries. Omni-Seg \cite{deng2023omni} uses an ensemble of CNN models trained on multiple image types to generalize across a wide range of biological imaging modalities without retraining. It applies domain adaptation and multi-task learning techniques. NuSeT \cite{yang2020nuset} is a CNN model tailored specifically for nuclear segmentation in fluorescence microscopy images, combining U-Net structure with a tailored loss function to enhance segmentation accuracy in noisy conditions.
\vspace{-1em}
\paragraph{3D Segmentation Methods.} We provide an overview of existing 3D segmentation techniques, along with their limitations in the context of the current state-of-the-art methods.

3DCellSeg \cite{wang2022novel} is a two-stage deep-learning pipeline designed for dense 3D cell segmentation in membrane-stained images. The pipeline is robust with only one main hyperparameter, avoiding extensive manual tuning and generalizing across different datasets and cell shapes. But still, as a supervised 3D-based method, 3DCellSeg still requires annotated training data for new cell types or imaging conditions. Its performance can degrade if applied to modalities it wasn’t trained on, necessitating retraining. Additionally, like many 3D-based methods, processing very large volumetric images can be computationally intensive comparing to 2D-based methods.

CellSeg3D \cite{achard2024cellseg3d} introduces a self-supervised 3D cell segmentation approach that eliminates the need for manual annotations. It employs a WNet3D architecture—comprising two 3D U-Nets connected sequentially—to perform segmentation by optimizing a soft normalized cuts loss directly on raw image volumes. However, as a self-supervised method, it relies on inherent image features, such as brightness differences, to delineate cells. This reliance may limit its ability to generalize effectively to end-user datasets, which often exhibit lower quality. While the performance is well on colorful fluorescent images, it may not be suitable for application in our generalized setting.

StarDist3D \cite{weigert2020stardist3d} is an instance segmentation method that represents cells (typically for nuclei) as star-convex shapes. A CNN predicts, for each pixel/voxel, the distances to the object boundary along a fixed set of radial rays, as well as an object probability score. StarDist assumes objects are approximately star-convex, so it may struggle with cells that have extremely irregular or concave shapes not well represented by a single star-shaped model. Another practical limitation is the need for training data for each new application. Due to the lack of large 3D training datasets, currently has no specific pretrained 3D StarDist models are publicly available. Still, StarDist3D achieves good performance on its specific cell types.

Go-Nuclear \cite{vijayan2024deep} is a recently introduced toolkit for 3D nucleus segmentation in tissue and organ datasets. Rather than a single algorithm, it encompasses a pipeline for generating training data and iteratively training models using multiple algorithms. The large curated dataset is used to train and fine-tune popular segmentation frameworks (CellPose, PlantSeg, StarDist) for nuclei in diverse organs. Yet, Go-Nuclear models are specialized for nuclear segmentation and rely on having a clearly labeled nucleus channel.

Due to the specificity of certain cell types and the lack of generalizability, as well as the absence of documentation on public open benchmarks, our work chooses to focus on CellPose, CellStitch, and PlantSeg, which have documented performance and are applicable to a wide range of cell types.
\vspace{-1em}
\paragraph{Geometry-aware Segmentation Methods.} We further introduce several geometry- and topology-aware segmentation methods. Hu et al. \citep{hu2019topology,hu2021topology} propose loss functions and training schemes that encourage a segmentation network to preserve topological structures during end-to-end training from raw images. \citet{lux2025topographefficientgraphbasedframework} define a topology-preserving loss framework based on a component graph for the joint ground-truth/prediction topology, again used during network training. \citet{pmlr-v202-stucki23a} introduce Betti matching, a spatially accurate loss for segmentation, by computing induced matchings between the persistence barcodes of prediction and ground truth. \citet{Shit_2021} propose centerline Dice, a similarity measure based on the intersection of segmentation masks with their skeletons, designed for tubular network-like structures, which is beyond the anisotropic 3D cellular segmentation studied in our work. 

In other words, existing geometry-aware methods primarily address how to \emph{train} a segmentation model, rather than how to perform a specific post-processing step on the output of such a model. These approaches are therefore complementary, not competing, to ours. All of them focus on loss functions for training a segmentation network from scratch or for fine-tuning it; their output is the segmentation itself. Moreover, these works mainly preserve \textbf{global topology} of one or a few foreground classes, whereas cell oversegmentation is an \textbf{instance-level} phenomenon. Frequently, Betti numbers and global connectivity are unchanged whether a single cell is split into two labels or kept as one, so these losses are blind to the specific oversegmentation errors that occur in a post-hoc manner. Adapting the ideas from the above methods to our setting would require substantial re-design to make them suitable as independent post-processing procedures for cellular segmentation. Therefore, we choose the most widely-used, state-of-the-art baselines and validate whether our method can further refine the segmentation quality over these methods. We hope that our framework will inspire follow-up work to adapt geometry-aware losses from prior segmentation methods into post-hoc processing methods.

\section{Technical Details for Implementation}
\subsection{Cross-layer Interpolation to Recover 2D Mis-segmented Cell Masks} 

As our pre-trained classifier finds the oversegmented candidates, we then need to recover the missing mask between the two oversegmented cells. To achieve this, we adopt the interpolation method proposed in CellStitch~\cite{Liu2023Cellstitch}, which builds upon the Wasserstein interpolation framework introduced by~\citet{Solomon2015}.

\begin{figure}[t]
    \centering
\includegraphics[width=0.55\textwidth]{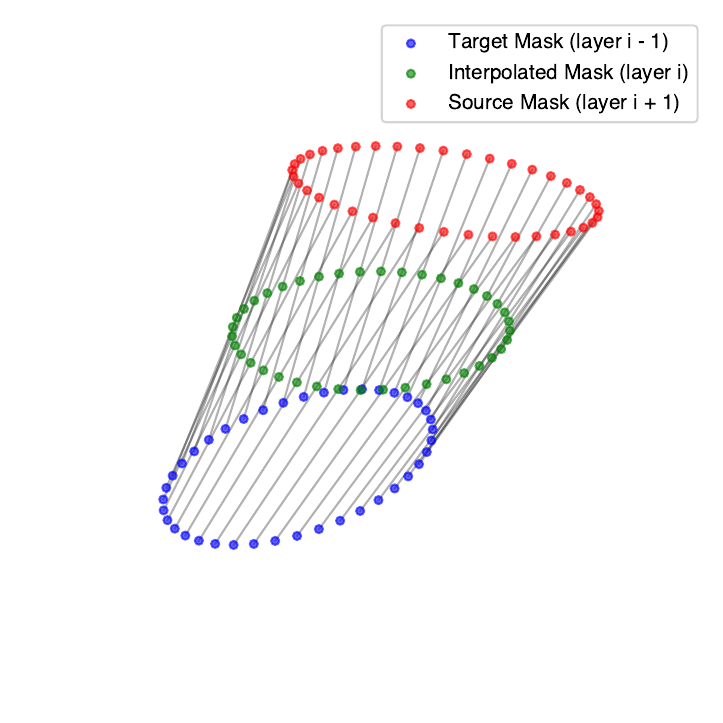} 
\vspace{-5em}
    \caption{Example of interpolation between the source mask at layer $i+1$ and the target mask at layer $i-1$.}\label{f7}
    \vspace{-1em}
\end{figure}

We begin by treating each matched pair of cells in consecutive slices as source and target boundaries (i.e., the contours of each cell mask). Each boundary pixel is assigned a uniform mass, and an optimal transport (OT) plan is computed to map source pixels to target pixels. For any intermediate layer, we interpolate each matched pixel pair based on the transport weights, resulting in a geometry-aware interpolation of the boundary. The interpolated boundaries are then filled to reconstruct the full cell mask at that layer. An example is illustrated in Figure~\hyperref[f7]{7}, where a mask at the intermediate layer~$i$ is generated using the cell contours from layers~$i+1$ and~$i-1$.

\subsection{Computational Efficiency}

We would like to further highlight our strong applicability in terms of computation efficiency. Since the number of cells, the average size of the cells, and the density of the cells varied from datasets to datasets, we believe providing a theoretical analysis towards the computational complexity of the algorithm is impractical. Therefore, we provide direct empirical results from real applications:

\begin{itemize}
    \item Processing a large $300 \times 2008 \times 2008 \ (\mathrm{Z} \times \mathrm{X} \times \mathrm{Y})$ animal stack takes approximately 1 hour in total on 2 NVIDIA A40 GPUs with 48GB storage.
    \item Running on a medium $2000 \times 224 \times 224  \ (\mathrm{Z} \times \mathrm{X} \times \mathrm{Y})$ plant dataset takes around 7 minutes to identify oversegmentation candidates and around 15 minutes to stitch and recover all oversegmented cells—using \textbf{CPUs} only.
\end{itemize}
\
This speed is primarily due to:

\begin{itemize}
    \item The use of sliced Wasserstein distance for EMD computation, which is well-suited to the uniform distribution of large cell masks, speeding up the calculation through random projections. 
    \item The lightweight structure of our pre-trained MLP model.
\end{itemize}

\paragraph{MLP Hyperparameters and Structure.} Our MLP is indeed lightweight and computationally efficient, which has 3 hidden layers with sizes 128, 64, and 32, uses ReLU activations, 0.3 dropout, a final sigmoid output.

\subsection{Dealing with Multi-Oversegmentations}

Although the occurrence of multiple gaps within a single cell is rare, we want to highlight that our framework is capable of handling such cases.

For example, consider a long cell body that is broken into multiple parts—say A, B, and C from top to bottom. Since our algorithm processes gaps layer by layer from top to bottom, it will first determine whether to stitch A and B. If A and B are stitched into a new cell body D, the algorithm will then assess whether to stitch D with C. This iterative process allows the framework to handle multiple fragments in a structured and consistent manner.
\begin{figure}[h]
    \centering
\includegraphics[width=0.8\textwidth]{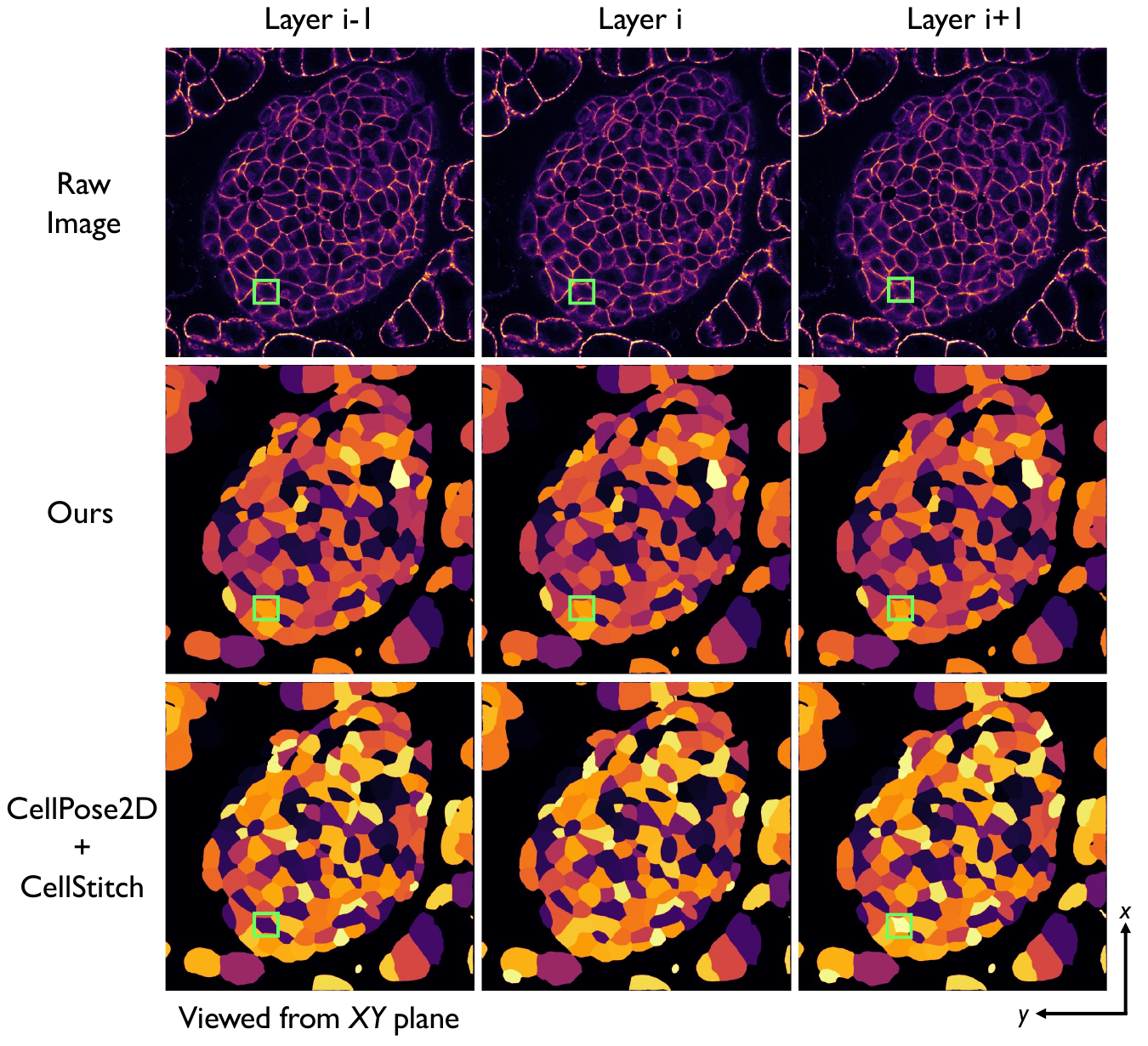} 
    \caption{Successful correction for the undersegmented 2D mask from the animal dataset, viewed from the $\mathrm{XY}$ plane. The positions of the mis-segmented masks are highlighted with green boxes.}
    \label{f8}
    \vspace{-1em}
\end{figure}
\section{Visualization Examples with Further Analysis}
In this section, we present several examples of correction results produced by our framework across various datasets and viewing planes to better illustrate its effectiveness. We begin by showcasing successful correction cases, followed by a closer examination of the unsure and incorrect cases. These examples also help us highlight the limitations of deterministic metrics such as mAP and Jaccard index, which may fail to fully capture the correctness of segmentation in complex scenarios.

\subsection{More Successful Examples}
Figure~\hyperref[f8]{8} illustrates an example of correcting a mis-segmentation that is not caused by an empty mask. In the CellStitch row, the purple cell at layer \( i-1 \) and the bright yellow cell at layer \( i+1 \) remain disconnected due to an undersegmentation error produced by CellPose2D at the same position in layer \( i \). Our method successfully identifies this oversegmentation from a 3D perspective and corrects the 2D undersegmentation. The raw image clearly indicates that the three cell masks belong to the same 3D cell body. Moreover, in line 6 of Algorithm~\hyperref[a1]{1}, we mean that there is no ``complete" cell existing between the gap, with both its highest and lowest layers located between our candidates. However, we allow noisy cell masks to exist between the candidates, enabling us to also address 2D undersegmentation errors.

\begin{figure}[t]
    \centering
\includegraphics[width=0.85\textwidth]{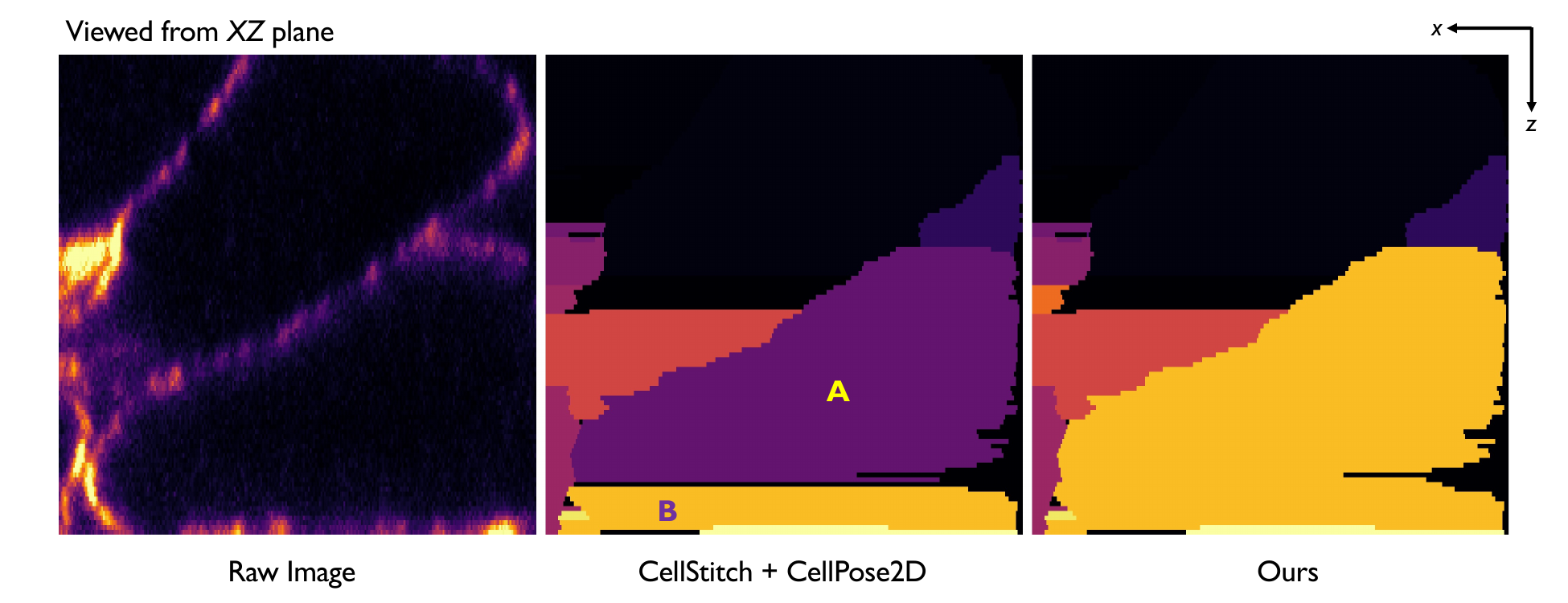} 
    \caption{Example of a successful correction in animal cell datasets, viewed from the $\mathrm{XZ}$ plane. The left panel displays the raw microscopic image, the middle panel shows the 3D segmentation produced by CellStitch, and the right panel presents our results, where Cell A (purple) and Cell B (yellow) from the middle panel were stitched.}
    \label{f9}
    \vspace{-1em}
\end{figure}

Figure~\hyperref[f9]{9} provides an example of correction viewed from the $\mathrm{XZ}$ plane, illustrating how two oversegmented cells (Cell A and Cell B) are stitched along the $\mathrm{Z}$ direction. Specifically, although the raw image shows that the cell's 3D topological shape follows a strictly circular pattern (a gradual monotonic increase followed by a decrease), 2D segmentation inaccuracies produced by CellPose2D introduce \textbf{oscillations} that deviate its shape from the strictly monotonic standard. However, our relaxation approach using regression \( \mathcal{R}^2 \) alleviates this issue, as small oscillations still yield a high \( \mathcal{R}^2 \) value, distinguishing it from natural gap cases. 

Figure~\hyperref[f10]{10} shows an example of accurate correction in the plant cell dataset by stitching the highlighted cell masks from layer \( i-1 \) and layer \( i+1 \). This correction is similar to the type shown in Figure~\hyperref[f8]{8}, where we also addressed 2D undersegmentation. The true 2D segmentation in layer \( i \) is recovered through cross-layer interpolation:

\begin{figure}[h]
    \centering
\includegraphics[width=0.75\textwidth]{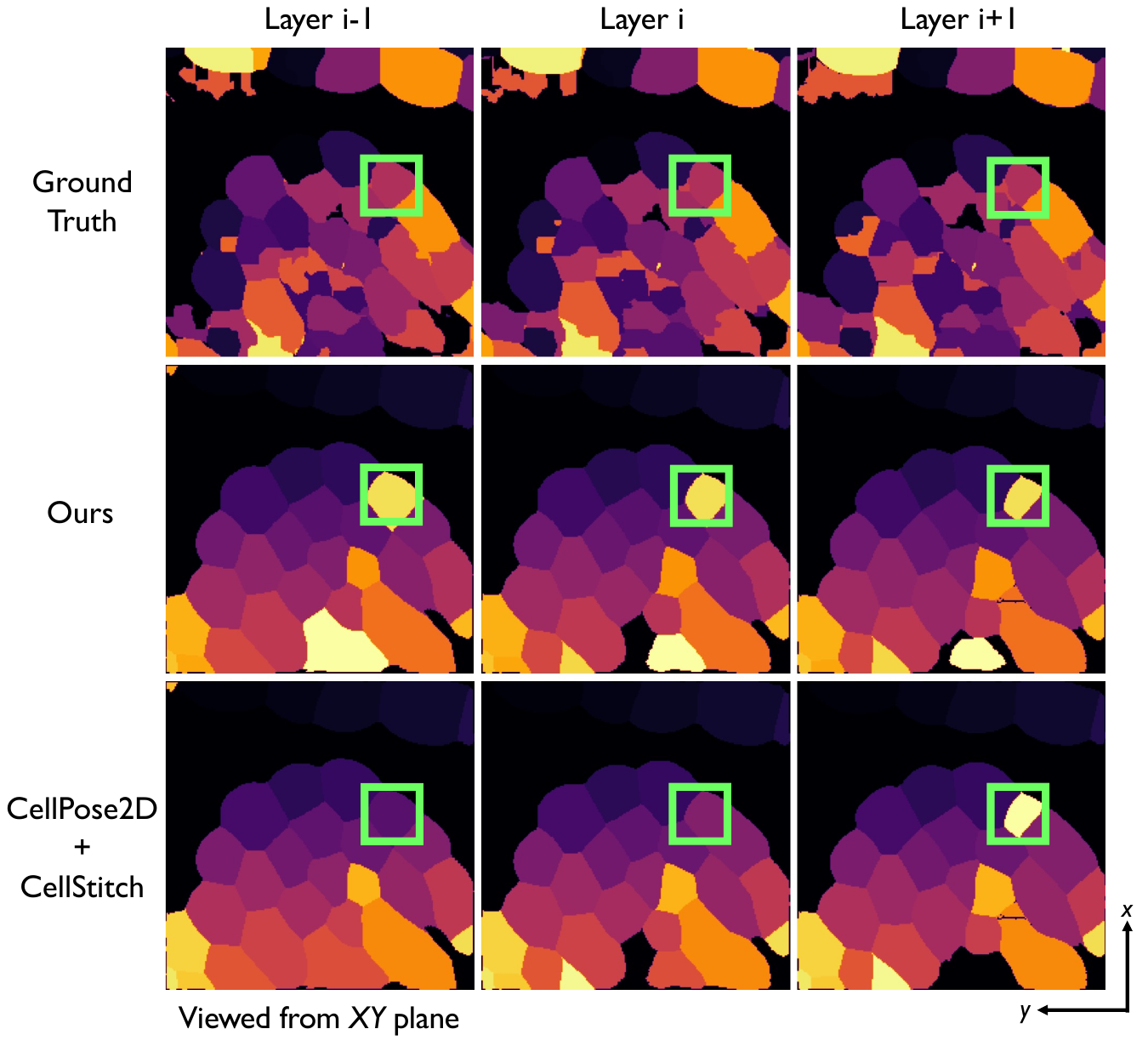} 
    \caption{Accurate correction viewed from XY plane from the plant cell dataset. Position of the mis-segmented masks are highlighted.}
    \label{f10}
\end{figure}

\subsection{Examples for Unsure Case}\label{ac3}
\begin{figure}[h]
    \centering
\includegraphics[width=0.75\textwidth]{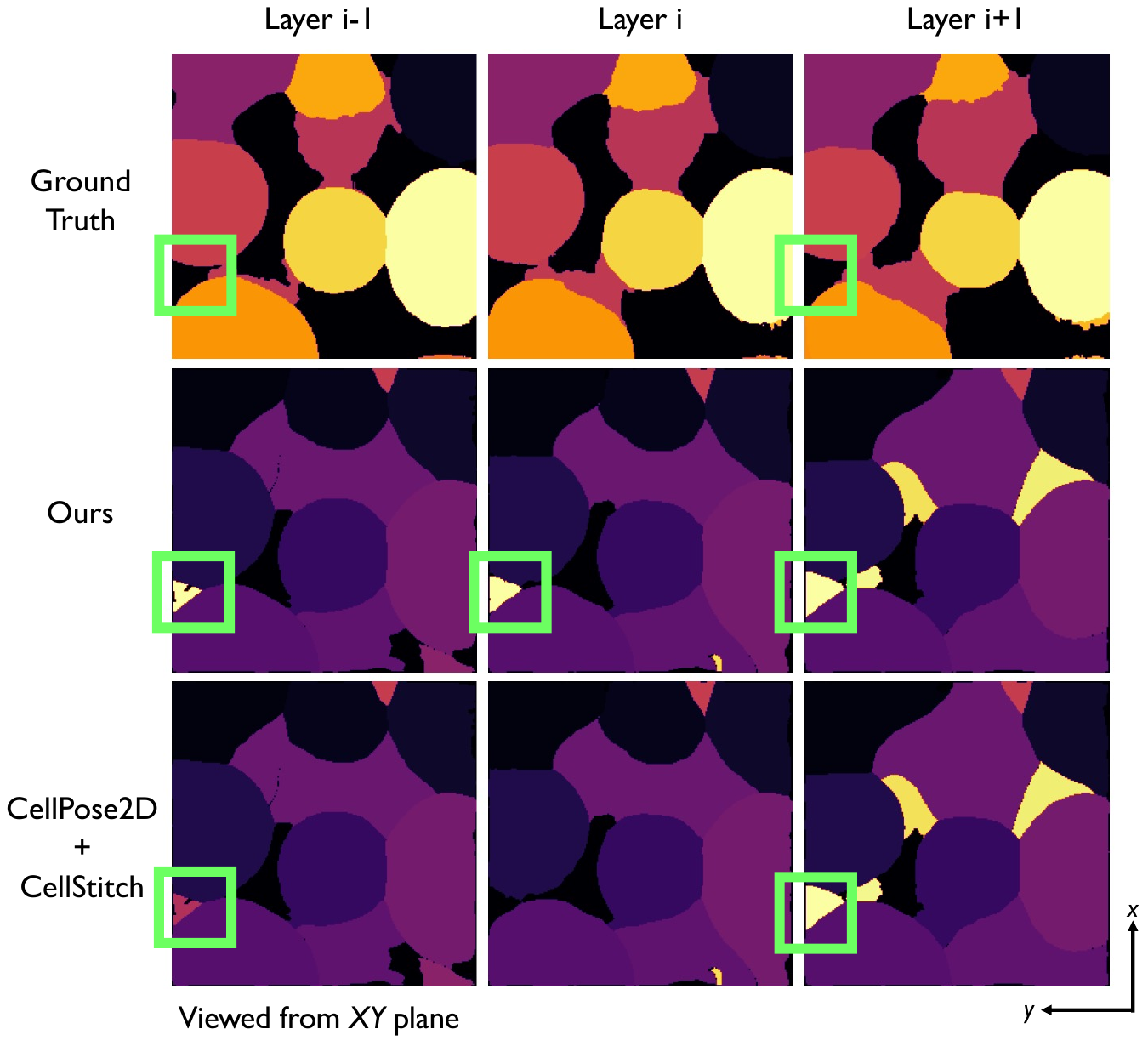} 
    \caption{Example of the unsure case from plant dataset.}
    \label{f13}
\end{figure}

\hyperref[f13]{Figure 11} presents an example of an unsure case caused by hallucination masks produced by CellPose2D. In the first row, the green box in the ground truth highlights an area where no cells are present. However, in the same position in the CellStitch result, two hallucinated masks (purple and yellow) are generated at layer \( i-1 \) and layer \( i+1 \). These noisy masks lead our framework to stitch them together. While the correction appears accurate, there is no evidence from the ground truth segmentation to support our judgment. Therefore, we classify these cases as unsure, \textbf{pessimistically reporting only the absolutely correct cases}.

\begin{figure}[t]
    \centering
\includegraphics[width=0.75\textwidth]{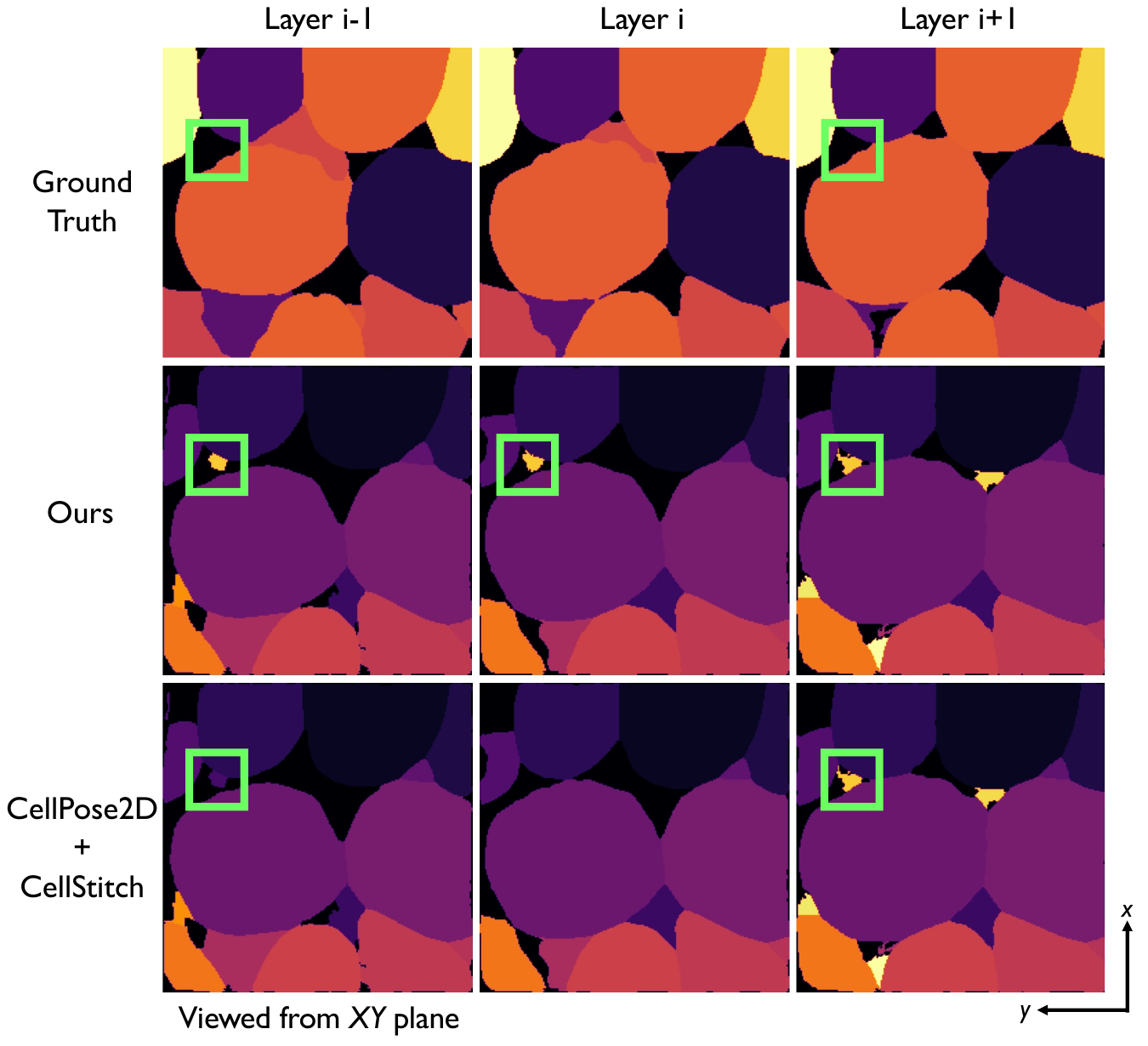} 

    \caption{Extra example of unsure case led by the noisy masks.}
    \label{f14}
\end{figure}

Figure~\hyperref[f14]{12} shows another uncertain case caused by noisy masks from CellPose2D. The green box in the ground truth marks an area without cells, but the CellStitch result includes two noisy masks (purple and orange) at layers \(i-1\) and \(i+1\), which are stitched together in our result. While the correction seems accurate, the ground truth provides no evidence to confirm this, so again, we conservatively report only the fully verified cases.

\subsection{Analysis on Incorrect Case}\label{ac4}
We also present example of incorrect cases to better understand the underlying issues in these situations and to provide clearer explanations for the causes of such incorrect results. As shown in Figure~\hyperref[f15]{14}, in the row of CellStitch results, the dark blue mask at layer \( i-1 \) and the yellow mask at layer \( i+1 \) are stitched together in our results. Without the ground truth labels, this stitching appears correct, as both cells are similar in shape and located in the same position, with a missing mask in the intermediate layer. However, the ground truth row reveals that the dark blue mask is part of the large purple cell in layer \( i-1 \), while the yellow mask belongs to the large orange cell in layer \( i+1 \). Thus, the dark blue and yellow masks actually belong to two different cells, and their stitching is incorrect, making this an example of an incorrect case.

\begin{figure}[h]
    \centering
\includegraphics[width=0.4\textwidth]{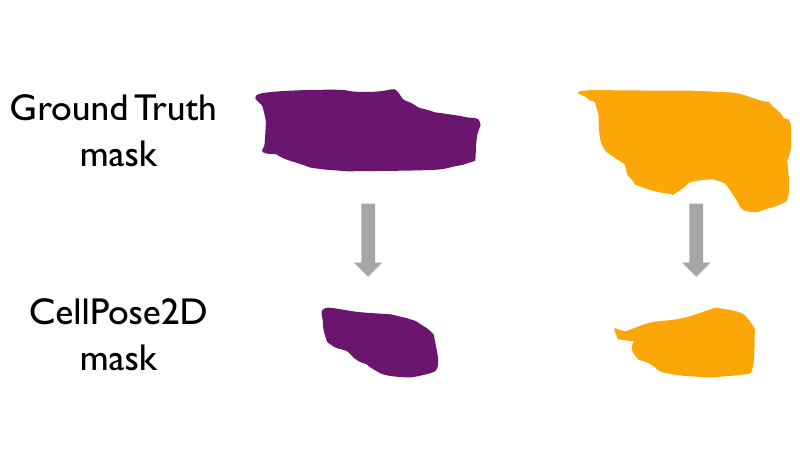} 
    \caption{Analysis towards the misleading shrunken masks shown in Figure~\hyperref[f15]{14}, where the cell masks are viewed from the XY plane.}
    \label{f16}
\end{figure}

\begin{figure}[t]
    \centering
\includegraphics[width=0.75\textwidth]{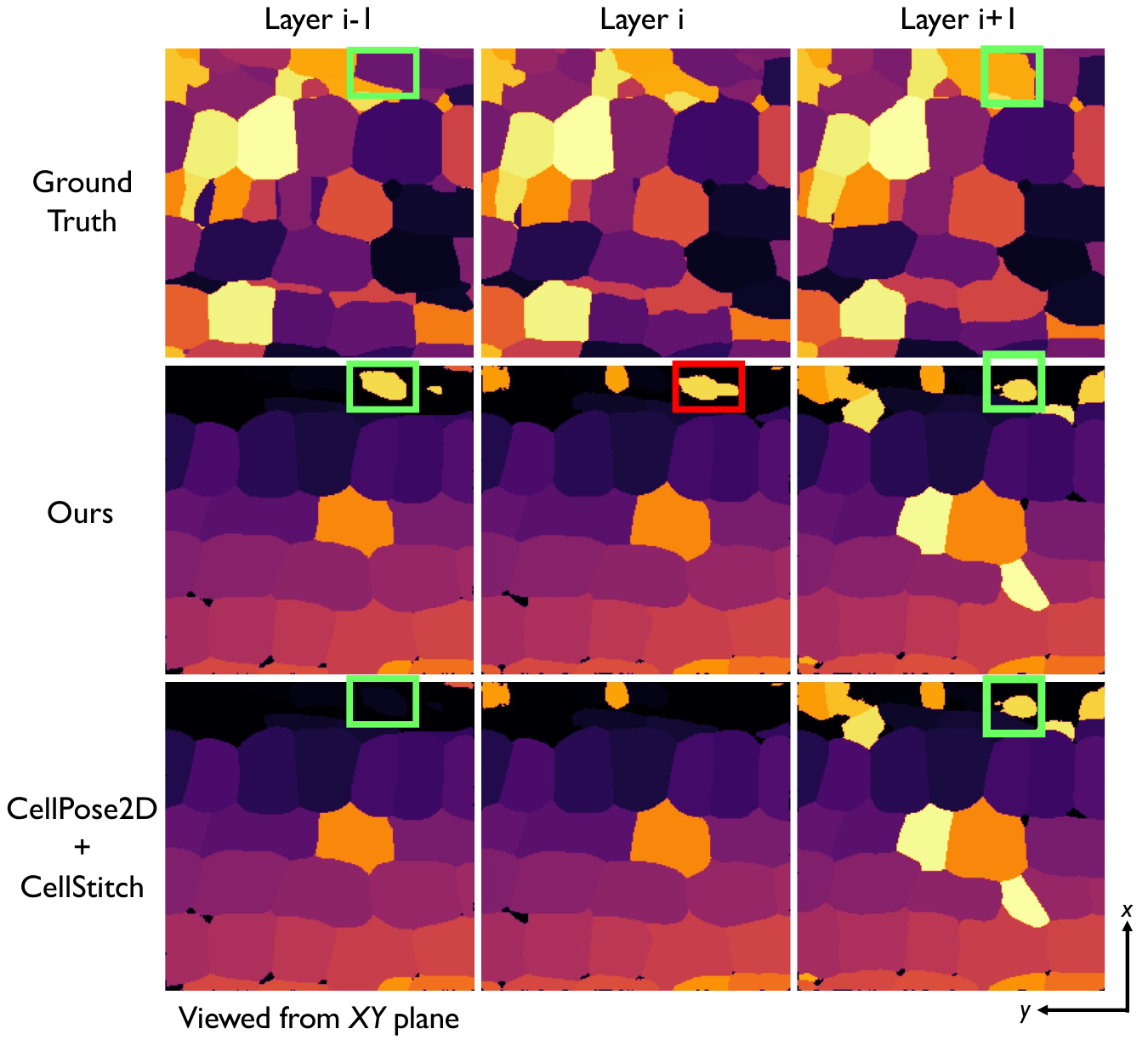} 
    \caption{An example of incorrect stitching, where two masks from different cells are erroneously stitched together. The mask in red box is highlighted as the incorrect recovered mask. The incorrect stitching is partially influenced by misleading information from the 2D segmentation results.}
    \label{f15}
\end{figure}

It is evident that both the dark blue and yellow masks only partially represent their corresponding ground truth cell masks. This highlights why we state that CellPose2D often ``shrinks the cell masks smaller" than they are supposed to be. We illustrate this effect in Figure~\hyperref[f16]{13}. When mask information is lost, both the geometric EMD measurement and the topological shape index are affected. For the EMD measurement, recall that EMD quantifies the effort required to transform one distribution into another. Losing geometric information alters the transformation, making the EMD between mask A and mask B differ from its original value. In the case shown in Figure~\hyperref[f16]{13}, the EMD between the mis-segmented masks is smaller compared to the ground truth masks, suggesting a falsely higher similarity. For the topological shape index, which is calculated via changes in overlapping areas, as shown in Figure~\hyperref[f15]{14}, the overlapping area between adjacent layers (e.g., layers \( i+1 \) and \( i-1 \)) aligns more closely with the overlapping areas of preceding and succeeding layers (e.g., layers \( i-2 \) and \( i+2 \)), as the masks shrink. These partial and altered representations lead our method to make incorrect judgments.

\subsection{Comparison between PlantSeg and CellStitch on Pancreas-B Dataset}
\begin{figure}[h!]
    \centering
\includegraphics[width=\textwidth]{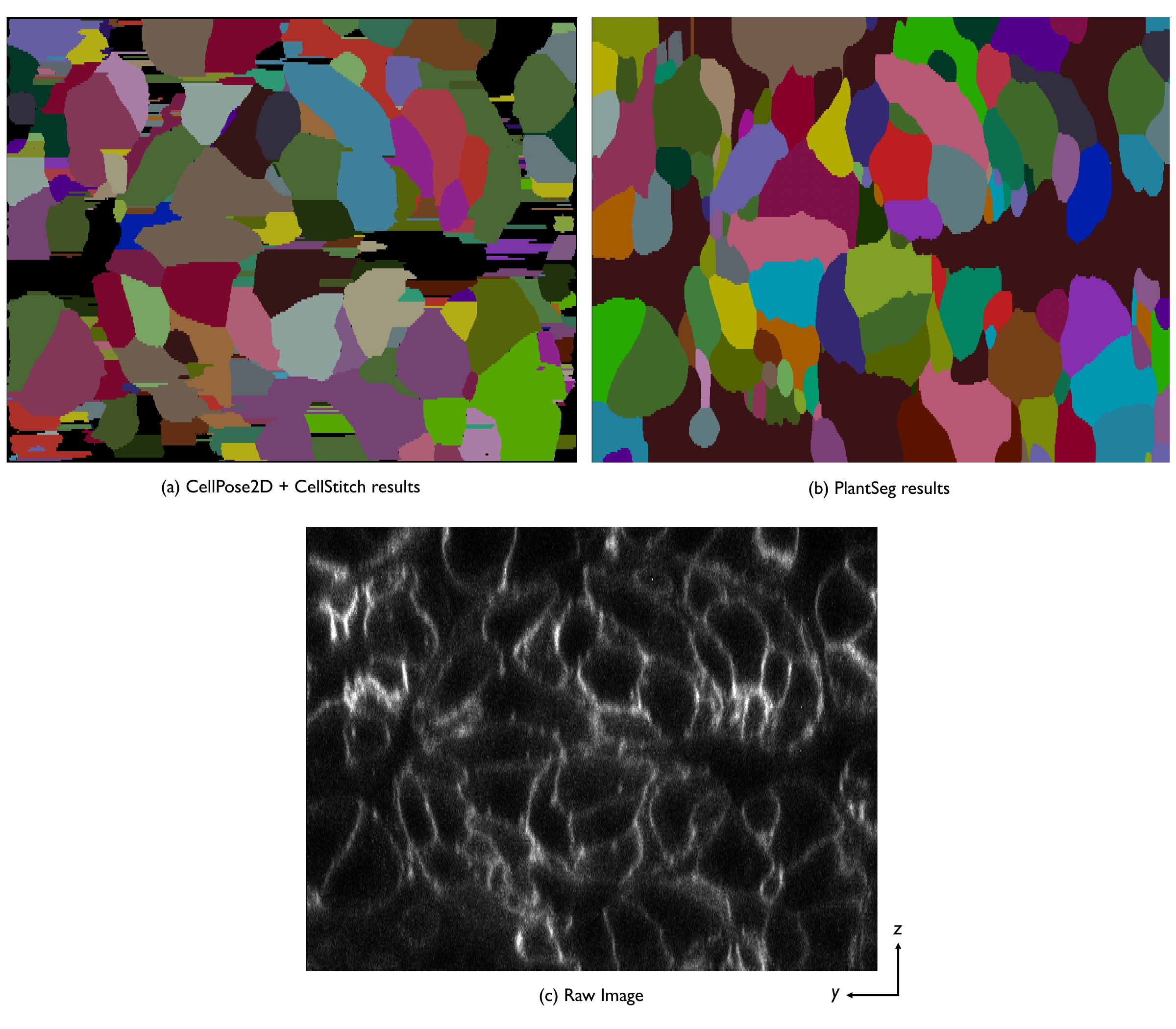} 
    \caption{Example from Pancreas-B dataset. (a) Final 3D segmentation results produced by CellStitch; (b) Final 3D segmentation results produced by PlantSeg; (c) Raw fluorescence image for cell membranes. A specific layer of 2D segmentation is selected and viewed from the YZ plane. Note that the images have been adjusted for $z$-anisotropy.}
    \label{f17}
\end{figure}

In Figure~\hyperref[f17]{15}, we show a representative $\mathrm{YZ}$-plane slice from the final 3D segmentation results. Most of the hallucinated noisy masks generated by CellPose2D are propagated into the final CellStitch output. Although PlantSeg also exhibits oversegmentation artifacts, its overall segmentation quality is better suited for end-users’ downstream analysis.

\section{Dataset Information} \label{ad}
\begin{table}[h]
\centering
\label{t7}
\resizebox{0.7\linewidth}{!}{%
\begin{tabular}{l|c|c|c|c}
\toprule
\textbf{Dataset} & \textbf{\# Stacks} & \textbf{Type} & \textbf{Labeled} & \textbf{Anisotropy ($z{:}y{:}x$)} \\
\midrule
Anther     & 100 & Plant / Public  & \checkmark & $4\!:\!1\!:\!1$ \\
Filament   & 100 & Plant / Public  & \checkmark & $4\!:\!1\!:\!1$ \\
Leaf       & 100 & Plant / Public  & \checkmark & $4\!:\!1\!:\!1$ \\
Pedicel    & 100 & Plant / Public  & \checkmark & $4\!:\!1\!:\!1$ \\
Sepal      & 100 & Plant / Public  & \checkmark & $4\!:\!1\!:\!1$ \\
Valve      & 100 & Plant / Public  & \checkmark & $4\!:\!1\!:\!1$ \\
Pancreas-A &  11 & Animal / Private & $\times$     & $4\!:\!1\!:\!1$ \\
Pancreas-B &   1 & Animal / Private & $\times$     & $4\!:\!1\!:\!1$ \\
\bottomrule
\end{tabular}
}
\caption{Overview of datasets.}
\end{table}

\hyperref[t7]{Table 5} shows the information for both plant and animal dataset. All the plant-type datasets are publicly availiable at \citet{bassel2019_dataset}. For plant-type data, the voxel resolution is unknown. For animal-type data, the voxel resolution ($\mathrm{Z} \times  \mathrm{X} \times \mathrm{Y} $) is $0.4 \times 0.1 \times 0.1~\text{\textmu m}$.

\paragraph{Pre-training Setup.} For CellPose, we use the pre-trained \texttt{cyto2} model to generate 2D pre-segmented results. We also tune the \texttt{diameter} case-by-case, from 60 to 100, based on the cells in each datasets. For PlantSeg, we use the pre-trained \texttt{generic\_confocal\_3D\_unet} model to generate 3D segmentation results.

\end{document}